\documentclass{article} 
\usepackage{iclr2019_conference,times}

\usepackage{amsmath,amsfonts,bm}

\def\eqref#1{equation~\ref{#1}}

\def\1{\bm{1}}

\DeclareMathAlphabet{\mathsfit}{\encodingdefault}{\sfdefault}{m}{sl}
\SetMathAlphabet{\mathsfit}{bold}{\encodingdefault}{\sfdefault}{bx}{n}

\usepackage[pagebackref=false,breaklinks=true,colorlinks,bookmarks=false,citecolor=blue]{hyperref}
\usepackage{url}
\usepackage{mathtools}
\usepackage{booktabs}
\usepackage{graphicx}
\usepackage{subcaption}
\usepackage{multirow}
\usepackage{algorithm}
\usepackage{algorithmic}

\title{InstaGAN:\\ Instance-aware Image-to-Image Translation}

\author{Sangwoo Mo$^*$, Minsu Cho$^\dagger$, Jinwoo Shin$^{*,\ddagger}$ \\
$^*$Korea Advanced Institute of Science and Technology (KAIST), Daejeon, Korea \\
$^\dagger$Pohang University of Science and Technology (POSTECH), Pohang, Korea \\
$^\ddagger$AItrics, Seoul, Korea \\
$^*$\texttt{\{swmo, jinwoos\}@kaist.ac.kr}, \quad $^\dagger$\texttt{mscho@postech.ac.kr}}

\DeclarePairedDelimiter\norm{\lVert}{\rVert}

\iclrfinalcopy 
\begin{document}

\maketitle

\vspace{-0.2in}
\begin{abstract}
\vspace{-0.05in}
Unsupervised image-to-image translation 
has gained considerable attention due to  
the recent impressive progress based on generative adversarial networks (GANs). However, previous methods often fail in challenging cases,
in particular, when an image has \textit{multiple} target instances and a translation task involves significant changes in \textit{shape},  \textit{e.g.,} translating pants to skirts in fashion images. 
To tackle the issues, we propose a novel method, coined \textit{instance-aware GAN (InstaGAN)},
that incorporates the \textit{instance} information
{(\textit{e.g.,} object  segmentation masks)}
and improves    \textit{multi-instance transfiguration}.
The proposed method translates both an image and the corresponding set of instance attributes while maintaining the permutation invariance property of the instances. 
To this end, we introduce a context preserving loss that encourages 
the network to learn the identity function outside of target instances.  
We also propose a sequential mini-batch inference/training technique  
that handles multiple instances with a limited GPU memory 
and enhances the network 
to generalize better for multiple instances. 
Our comparative evaluation demonstrates the effectiveness of the proposed method on different image datasets, in particular, in the aforementioned challenging cases.
Code and results are available in \url{https://github.com/sangwoomo/instagan}.
\end{abstract}
\vspace{-0.05in}

\vspace{-0.05in}
\section{Introduction}
\label{sec:intro}
\vspace{-0.05in}

\begin{figure}[t]
\vspace{-0.25in}
\centering
\includegraphics[width=\textwidth]{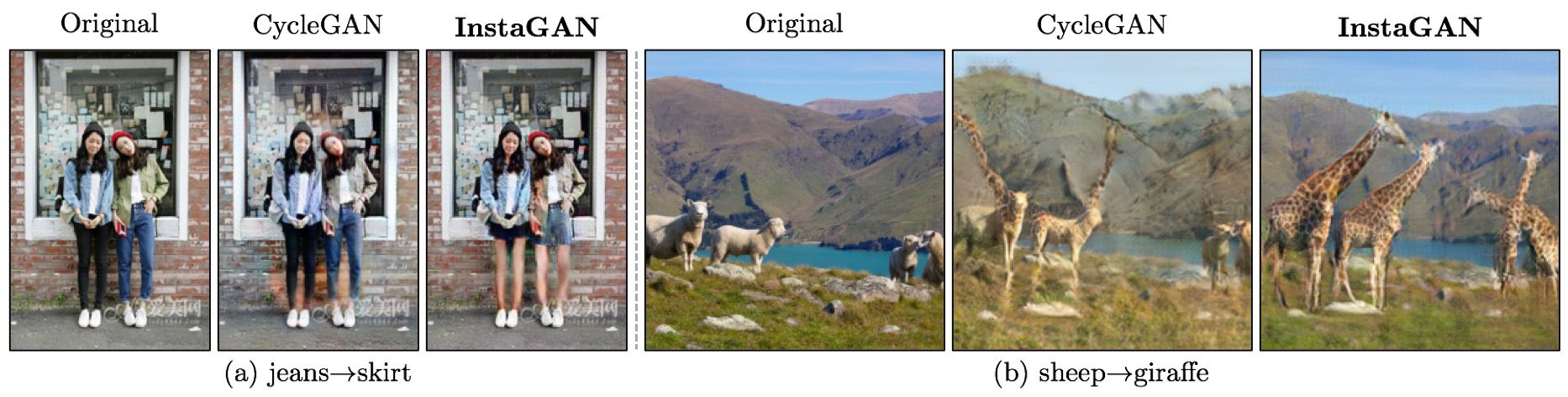}
\caption{
Translation results of the prior work (CycleGAN, \citet{zhu2017unpaired}), and our proposed method, InstaGAN.
Our method shows better results for multi-instance transfiguration problems.
} \label{fig:motivation}
\vspace{-0.15in}
\end{figure}

Cross-domain generation arises in many machine learning tasks,
including neural machine translation \citep{artetxe2017unsupervised, lample2017unsupervised},  
image synthesis \citep{reed2016generative, zhu2016generative},
text style transfer \citep{shen2017style},
and video generation \citep{bansal2018recycle, wang2018video, chan2018everybody}.
In particular, the unpaired (or unsupervised) image-to-image translation has achieved an impressive progress based on variants of generative adversarial networks (GANs) 
\citep{zhu2017unpaired, liu2017unsupervised, choi2017stargan, almahairi2018augmented, huang2018multimodal, lee2018diverse}, and has also drawn considerable attention due to its practical applications including 
colorization \citep{zhang2016colorful}, super-resolution \citep{ledig2017photo}, semantic manipulation \citep{wang2018high},
and domain adaptation \citep{bousmalis2017unsupervised, shrivastava2017learning, hoffman2017cycada}.
Previous methods on this line of research, however, 
often fail on challenging tasks,
in particular, when the translation task involves significant changes in  \textit{shape} of instances  \citep{zhu2017unpaired}
or the images to translate contains \textit{multiple} target instances \citep{gokaslan2018improving}.
Our goal is to extend    image-to-image translation towards such challenging tasks,
which can strengthen its applicability up to the next level,
\textit{e.g.,} changing pants to skirts in fashion images for a customer to decide which one is better to buy.
To this end, we propose a novel method that incorporates the \textit{instance} information of multiple target objects
in the framework of generative adversarial networks (GAN);  hence we called it {\it instance-aware GAN (InstaGAN)}.
In this work, we use the object segmentation masks for instance information,
which may be a good representation for instance shapes, as it contains object boundaries while ignoring other details such as color.
Using the information, our method shows impressive results for
\textit{multi-instance transfiguration} tasks, as shown in Figure \ref{fig:motivation}.

Our main contribution is three-fold: an instance-augmented neural  architecture, a context preserving loss,
and a sequential mini-batch inference/training technique. 
First, we propose a neural network architecture that  translates both an image and the corresponding set of instance attributes. 
Our architecture can translate an arbitrary number of instance attributes conditioned by the input,
and is designed to be permutation-invariant to the order of instances.
Second, we propose a context preserving loss that 
encourages the network to focus on target instances in translation and learn an identity function outside of them. 
Namely, it aims at preserving the background context while transforming the target instances.
Finally, we propose a sequential mini-batch inference/training technique,
\textit{i.e.,} translating the mini-batches of instance attributes sequentially, instead of doing the entire set at once.
It allows to handle a large number of instance attributes with a limited GPU memory,
and thus enhances the network to generalize better for images with many instances.
Furthermore, it improves the translation quality of images with even a few instances because it acts as data augmentation during training by producing  multiple intermediate samples. 
All the aforementioned contributions are dedicated to how to
incorporates the instance information (\textit{e.g.,} segmentation masks) for image-to-image translation.
However, we believe that our approach is applicable to numerous other cross-domain generation tasks where set-structured side information is available.


To the best of our knowledge, 
we are the first to report image-to-image translation results for multi-instance transfiguration tasks.
A few number of recent methods  \citep{kim2017learning, liu2017unsupervised, gokaslan2018improving} 
show some transfiguration results 
but only for images with a single instance often in a clear background. 
Unlike the previous results in a simple setting, our focus is on the \textit{harmony} of instances naturally rendered with the background.
On the other hand, CycleGAN \citep{zhu2017unpaired} show
some results for multi-instance cases,
but report only a limited performance for transfiguration tasks. 
At a high level, the significance of our work is also on 
discovering that the instance information is effective for shape-transforming image-to-image translation, which we think would be influential to other related research in the future.
Mask contrast-GAN \citep{liang2017generative} and
Attention-GAN \citep{mejjati2018unsupervised} use segmentation masks or predicted attentions,
but only to attach the background to the (translated) cropped instances. They do not allow to transform the shapes of the instances.
To the contrary, our method learns how to preserve the background by optimizing 
the context preserving loss, thus facilitating the shape transformation.

\vspace{-0.05in}
\section{InstaGAN: Instance-aware Image-to-Image Translation}
\label{sec:method}
\vspace{-0.05in}

Given two image domains $\mathcal{X}$ and $\mathcal{Y}$, the problem of image-to-image translation aims to learn
mappings across different image domains, $G_\textrm{XY}: \mathcal{X} \to \mathcal{Y}$ or/and $G_\textrm{YX}: \mathcal{Y} \to \mathcal{X}$, \textit{i.e.,} transforming target scene elements while preserving the {original contexts.} 
This can also be formulated as a conditional generative modeling task where we estimate the conditionals $p(y|x)$ or/and $p(x|y)$. 
The goal of {\it unsupervised} translation we tackle is to recover such mappings only using unpaired
samples from marginal distributions of original data,
$p_{\tt data}(x)$ and $p_{\tt data}(y)$ of two image domains.

The main and unique idea of our approach is to incorporate the additional \textit{instance} information,
\textit{i.e.,} augment a space of \textit{set of instance attributes} $\mathcal{A}$ to the original image space $\mathcal{X}$,
to improve the image-to-image translation.
The set 
of instance attributes $\bm{a} \in \mathcal{A}$ comprises all individual attributes of $N$ target instances: $\bm{a} = \{a_i\}_{i=1}^{N}$. In this work, we use an instance segmentation mask only, but we remark that any useful type of instance information can be incorporated for the attributes. 
Our approach then can be described as learning joint-mappings between attribute-augmented spaces 
$\mathcal{X} \times \mathcal{A}$ and $\mathcal{Y} \times \mathcal{B}$.
This leads to disentangle different instances in the image
and allows the generator to perform an accurate and detailed translation.
We learn our attribute-augmented mapping in the framework of generative adversarial networks (GANs) \citep{goodfellow2014generative},
hence, we call it \textit{instance-aware GAN (InstaGAN)}. 
We present details of our approach in the following subsections. 


\vspace{-0.05in}
\subsection{InstaGAN Architecture}
\label{sec:method-architecture}
\vspace{-0.05in}

\begin{figure}[t]
\vspace{-0.2in}
\centering
\begin{subfigure}{.26\textwidth}
	\centering
	\includegraphics[height=5.6cm]{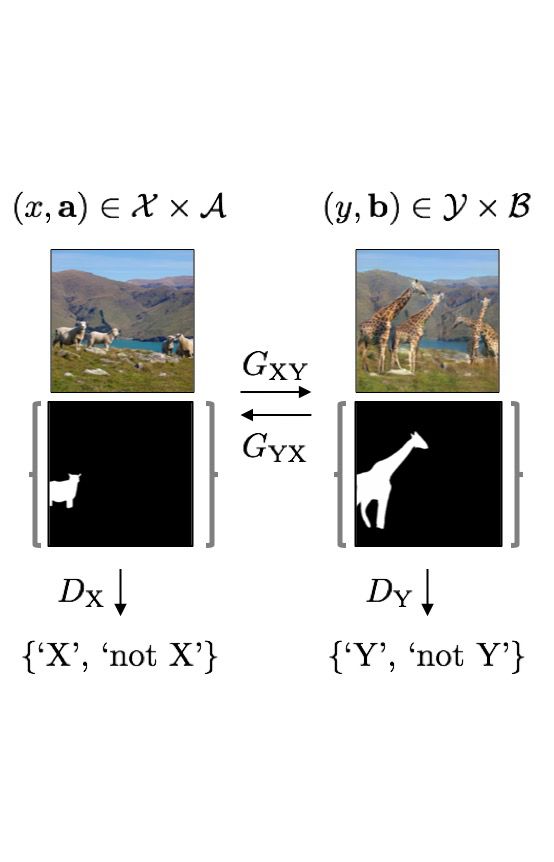}
	\caption{Overview}\label{fig:architecturea}
\end{subfigure}
\begin{subfigure}{.38\textwidth}
	\centering
	\includegraphics[height=5.6cm]{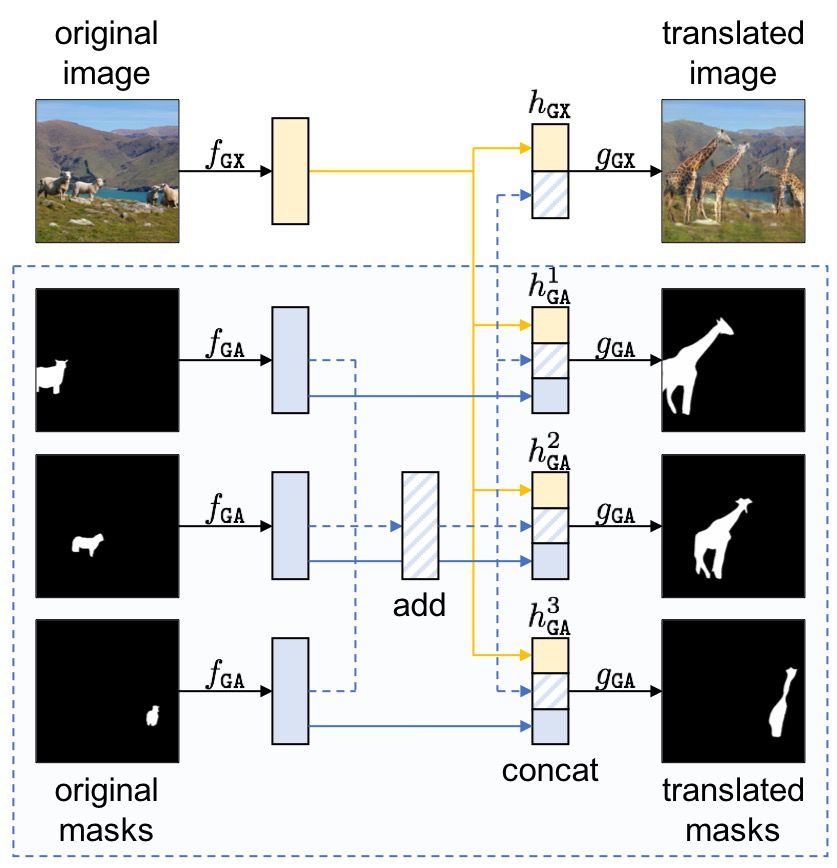}
	\caption{Generator $G$}\label{fig:architectureb}
\end{subfigure}
\begin{subfigure}{.34\textwidth}
	\centering
	\includegraphics[height=5.6cm]{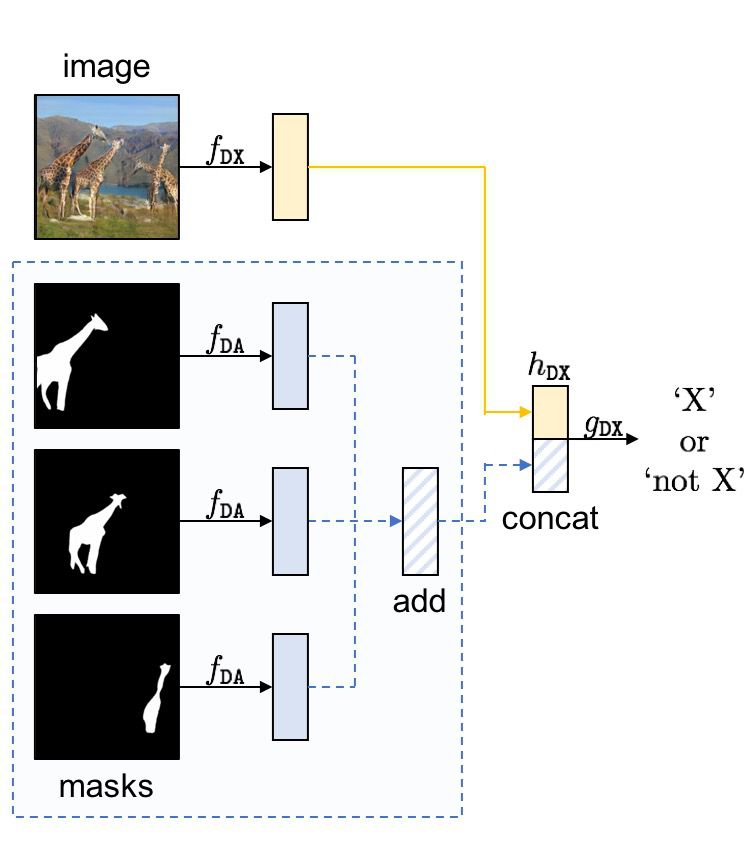}
	\caption{Discriminator $D$}\label{fig:architecturec}
\end{subfigure}
\caption{
(a) Overview of InstaGAN,
where generators $G_\textrm{XY}$, $G_\textrm{YX}$ and discriminator $D_\textrm{X}$, $D_\textrm{Y}$ follows the architectures in (b) and (c), respectively.
Each network is designed to encode both an image and set of instance masks.
$G$ is permutation equivariant, and $D$ is permutation invariant to the set order.
To achieve properties,
we sum features of all set elements for invariance, 
and then concatenate it with the identity mapping for equivariance. 
} \label{fig:architecture}
\vspace{-0.1in}
\end{figure}

Recent GAN-based methods~\citep{zhu2017unpaired, liu2017unsupervised} have achieved impressive performance in the unsupervised translation by jointly training two coupled mappings $G_\textrm{XY}$ and $G_\textrm{YX}$ with a cycle-consistency loss that encourages $G_\textrm{YX}(G_\textrm{XY}(x)) \approx x$ and $G_\textrm{XY}(G_\textrm{YX}(y)) \approx y$.
Namely, we choose to leverage the CycleGAN approach~\citep{zhu2017unpaired} to build our InstaGAN.
However, we remark that training two coupled mappings is not essential for our method,
and one can also design a single mapping following other approaches \citep{benaim2017one, galanti2018role}.
Figure \ref{fig:architecture} illustrates the overall architecture of our model.
We train two coupled generators $G_\textrm{XY}: \mathcal{X} \times \mathcal{A} \to \mathcal{Y} \times \mathcal{B}$
and $G_\textrm{YX}: \mathcal{Y} \times \mathcal{B} \to \mathcal{X} \times \mathcal{A}$,
where $G_\textrm{XY}$ translates the original data $(x,\bm{a})$ to
the target domain data $(y',\bm{b'})$ (and vice versa for $G_\textrm{YX}$),
with adversarial discriminators $D_\textrm{X}:\mathcal{X} \times \mathcal{A} \to \{\mbox{`X', `not X'}\}$
and $D_\textrm{Y}:\mathcal{Y} \times \mathcal{B} \to \{\mbox{`Y', `not Y'}\}$,
where $D_\textrm{X}$ determines if the data (original $(x,\bm{a})$ or translated $(x',\bm{a}')$)
is in the target domain $\mathcal{X} \times \mathcal{A}$ or not (and vice versa for $D_\textrm{Y}$).

Our generator $G$ encodes both $x$ and $\bm{a}$, and translates them into $y'$ and $\bm{b}'$.
Notably, the order of the instance attributes in the set $\bm{a}$ should not affect the translated image $y'$, and each instance attribute in the set $\bm{a}$ should be translated to the corresponding one in $\bm{b}'$. In other words, $y'$ is \textit{permutation-invariant} with respect to the instances in $\bm{a}$, and $\bm{b}'$  is \textit{permutation-equivariant} with respect to them. These properties can be implemented by introducing proper operators in feature encoding~\citep{zaheer2017deep}. 
We first extract \textit{individual} features from image and attributes using image feature extractor $f_{\tt GX}$ and attribute feature extractor $f_{\tt GA}$, respectively. The attribute features individually extracted using $f_{\tt GA}$ are then aggregated into a permutation-invariant set feature via summation: $\sum_{i=1}^N f_{\tt GA}(a_i)$. 
As illustrated in Figure~\ref{fig:architectureb}, we concatenate some of image and attribute features with the set feature, and feed them to image and attribute generators.
{Formally,
the image representation $h_{\tt GX}$ and the $n$-th attribute representation $h^n_{\tt GA}$ in generator $G$ can be formulated as: 
}
\begin{equation}
	h_{\tt GX}(x,\bm{a}) = \left[ f_{\tt GX}(x); \sum_{i=1}^N f_{\tt GA}(a_i) \right], \;\;\;
	h_{\tt GA}^n(x,\bm{a}) = \left[ f_{\tt GX}(x); \sum_{i=1}^N f_{\tt GA}(a_i); f_{\tt GA}(a_n) \right],
\end{equation}
where each attribute encoding $h_{\tt GA}^n$ process features of all attributes as a contextual feature.  
Finally, $h_{\tt GX}$ is fed to the image generator $g_{\tt GX}$, and $h_{\tt GA}^n$ $(n=1,\dots,N)$ are to the attribute generator $g_{\tt GA}$.

On the other hand, our discriminator $D$ encodes both $x$ and $\bm{a}$ {(or $x'$ and $\bm{a}'$)},
and determines whether the pair is from the domain or not. Here, the order of the instance attributes in the set $\bm{a}$ should not affect the output. 
In a similar manner above, our representation in discriminator $D$, which is permutation-invariant to the instances, is formulated as:
\begin{equation}
	h_{\tt DX}(x,\bm{a}) = \left[ f_{\tt DX}(x); \sum_{i=1}^N f_{\tt DA}(a_i) \right], 
\end{equation}
which is fed to an adversarial discriminator $g_{\tt DX}$. 

We emphasize that the joint encoding of both image $x$ and instance attributes $\bm{a}$ 
for each neural component is crucial because 
it allows the network to learn the \textit{relation} between $x$ and $\bm{a}$.
For example, if two separate encodings and discriminators are used for $x$ and $\bm{a}$,
the generator may be misled to produce image and instance masks that do not match with each other.
By using the joint encoding and discriminator, our generator can produce an image of
instances properly depicted on the area consistent with its segmentation masks.
As will be seen in Section \ref{sec:exp}, our approach can disentangle output instances considering their original layouts.
Note that any types of neural networks may be used for sub-network architectures mentioned above such as $f_{\tt GX}$, $f_{\tt GA}$, $f_{\tt DX}$, $f_{\tt DA}$, $g_{\tt GX}$, $g_{\tt GA}$, and $g_{\tt DX}$.
We describe the detailed architectures used in our experiments in Appendix \ref{sec:architecture-details}.


\vspace{-0.025in}
\subsection{Training Loss}
\label{sec:method-loss}
\vspace{-0.025in}

Remind that an image-to-image translation model aims to translate a domain
while keeping the original contexts (\textit{e.g.,} background or instances' domain-independent characteristics such as the looking direction).
To this end, we both consider the \textit{domain} loss, which makes the generated outputs to follow the style of a target domain,
and the \textit{content} loss, which makes the outputs to keep the original contents.
Following our baseline model, CycleGAN \citep{zhu2017unpaired},
we use the GAN loss for the domain loss, and consider both the cycle-consistency loss \citep{kim2017learning, yi2017dualgan}
and the identity mapping loss \citep{taigman2016unsupervised} for the content losses.\footnote{
We remark that the identity mapping loss is also used in CycleGAN (see Figure 9 of \citet{zhu2017unpaired}).
}
In addition, we also propose a new content loss, coined \textit{context preserving loss},
using the original and predicted segmentation information.
In what follows, we formally define our training loss in detail. 
For simplicity, we denote our loss function as a function of a single training sample
$(x, \bm{a}) \in \mathcal{X} \times \mathcal{A}$ and $(y, \bm{b}) \in \mathcal{Y} \times \mathcal{B}$,
while one has to minimize its empirical means in training.

The GAN loss is originally proposed by \citet{goodfellow2014generative}
for generative modeling via alternately training generator $G$ and discriminator $D$.
Here, $D$ determines if the data is a real one of a fake/generated/translated one made by $G$.
There are numerous variants of the GAN loss 
\citep{nowozin2016f, arjovsky2017wasserstein, li2017mmd, mroueh2017sobolev}, and 
we follow the LSGAN scheme \citep{mao2017least}, which is empirically known to show a stably good performance:
\begin{equation}
    \mathcal{L}_{\tt LSGAN} = 
	(D_\textrm{X}(x,\bm{a})-1)^2 + D_\textrm{X}(G_\textrm{YX}(y,\bm{b}))^2 + (D_\textrm{Y}(y,\bm{b})-1)^2 + D_\textrm{Y}(G_\textrm{XY}(x,\bm{a}))^2.
    \label{eq:gan}
\end{equation}
For keeping the original content,
the cycle-consistency loss $\mathcal{L}_{\tt cyc}$ and 
the identity mapping loss $\mathcal{L}_{\tt idt}$
enforce samples 
not to lose the original information
after translating twice and once, respectively:
\begin{align}
	\mathcal{L}_{\tt cyc}
	&= \norm{G_\textrm{YX}(G_\textrm{XY}(x,\bm{a}))-(x,\bm{a})}_1 + \norm{G_\textrm{XY}(G_\textrm{YX}(y,\bm{b}))-(y,\bm{b})}_1,
	\label{eq:cyc}\\
	\mathcal{L}_{\tt idt}
	&= \norm{G_\textrm{XY}(y,\bm{b})-(y,\bm{b})}_1 + \norm{G_\textrm{YX}(x,\bm{a})-(x,\bm{a})}_1.
	\label{eq:idt}
\end{align}
Finally, our newly proposed context preserving loss $\mathcal{L}_{\tt ctx}$
enforces to translate instances only, while keeping outside of them, \textit{i.e.,} background.
Formally, it is a pixel-wise weighted $\ell_1$-loss where the weight is
$1$ for background and $0$ for instances. 
Here, note that backgrounds for two domains become different in transfiguration-type translation involving significant shape changes. 
Hence, we consider the non-zero weight only if a pixel is in background
in both original and translated ones. 
Namely,  for the original samples $(x, \bm{a})$, $(y, \bm{b})$ and the translated one $(y', \bm{b}')$, $(x', \bm{a}')$, 
we let the weight $w(\bm{a},\bm{b}')$, $w(\bm{b},\bm{a}')$
be one minus
the element-wise minimum of binary represented instance masks,
and we propose
\begin{equation}
	\mathcal{L}_{\tt ctx}
	= \norm{w(\bm{a},\bm{b}') \odot (x - y')}_1] + \norm{w(\bm{b}, \bm{a}') \odot (y - x')}_1
	\label{eq:ctx}
\end{equation}
where $\odot$ is the element-wise product.
In our experiments,
we found that
the context preserving loss not only keeps the background better, but also improves the quality of generated instance segmentations.
Finally, the total loss of InstaGAN is
\begin{equation}
	\mathcal{L}_{\tt InstaGAN} = \underbrace{\mathcal{L}_{\tt LSGAN}}_{\tt GAN~(domain)~loss} + 
	\underbrace{\lambda_{\tt cyc} \mathcal{L}_{\tt cyc} + \lambda_{\tt idt} \mathcal{L}_{\tt idt}
	+ \lambda_{\tt ctx} \mathcal{L}_{\tt ctx}}_{\tt content~loss},
	\label{eq:loss}
\end{equation}
where
$\lambda_{\tt cyc}, \lambda_{\tt idt}, \lambda_{\tt ctx}>0$ are 
some hyper-parameters balancing the losses.

\vspace{-0.025in}
\subsection{Sequential Mini-Batch Translation}
\label{sec:method-algorithm}
\vspace{-0.025in}

\begin{figure}[t]
\vspace{-0.2in}
\centering
\includegraphics[width=\textwidth]{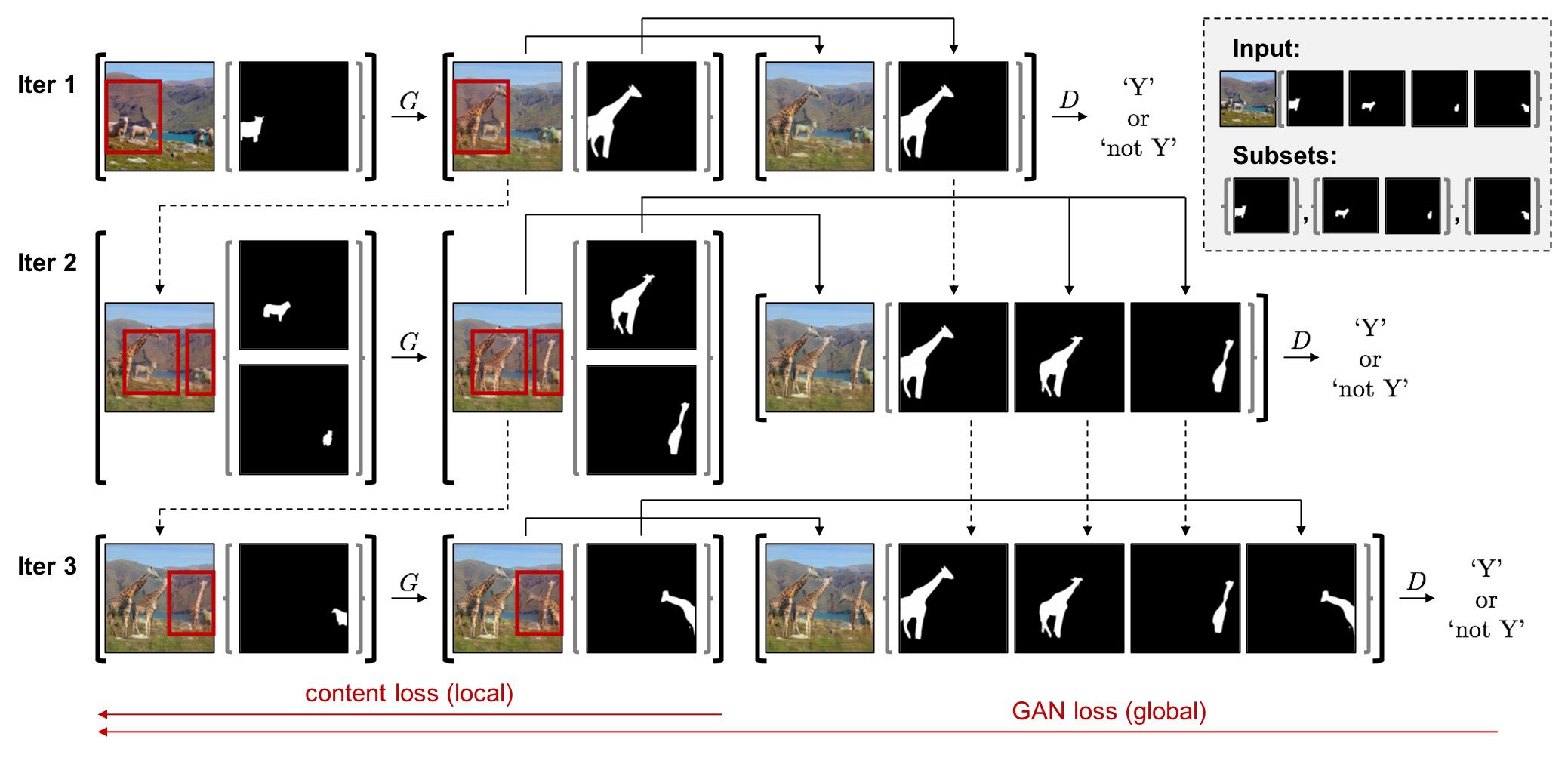}
\caption{
Overview of the sequential mini-batch training with instance subsets (mini-batches) of size 1,2, and 1, as shown in the top right side.
The content loss is applied to the intermediate samples of current mini-batch,
and GAN loss is applied to the samples of aggregated mini-batches.
We detach every iteration in training, in that
the real line indicates the backpropagated paths and dashed lines indicates the detached paths. See text for details. 
} \label{fig:sequential}
\vspace{-0.1in}
\end{figure}



While the proposed architecture is able to translate an arbitrary number of instances in principle,
the GPU memory required linearly increases with the number of instances. 
For example, in our experiments, a machine was able to forward only a small number (say, 2) of instance attributes  during training,
and thus the learned model suffered from poor generalization to images with a larger number of instances.
To address this issue, we propose a new inference/training technique, 
which allows to train an arbitrary number of instances without increasing the GPU memory.
We first describe the sequential inference scheme that
translates the subset of instances sequentially, and then describe the corresponding mini-batch training technique.

Given an input $(x, \bm{a})$,
we first divide the set of instance masks $\bm{a}$ into  mini-batches $\bm{a}_1, \dots, \bm{a}_M$, \textit{i.e.,} $\bm a = \bigcup_i \bm a_i$ and $\bm a_i \cap \bm a_j =\emptyset$ for $i\neq j$.
Then, at the $m$-th iteration for $m=1,2,\dots, M$, we translate the image-mask pair $(x_m, \bm{a}_m)$,
where $x_m$ is the 
translated image $y'_{m-1}$ from the previous iteration, and $x_1 = x$. 
In this sequential scheme, at each iteration, the generator $G$ outputs an intermediate translated image $y'_m$, which accumulates all mini-batch translations up to the current iteration, and a translated mini-batch of instance masks $\bm{b}'_{m}$:
\begin{equation}
    (y'_m, \bm{b}'_{m}) = G(x_m, \bm{a}_m) = G(y'_{m-1}, \bm{a}_m). 
	\label{eq:local}
\end{equation}
In order to align the translated  image with mini-batches of instance masks, we aggregate all the translated mini-batch and produce a translated sample:
\begin{equation}
	(y'_m, \bm{b}'_{1:m})
	= (y'_m, \cup_{i=1}^m \bm{b}'_i).
	\label{eq:global}
\end{equation}
The final output of the proposed sequential inference scheme is $(y'_M, \bm{b}'_{1:M})$.


We also propose the corresponding sequential training algorithm, as illustrated in Figure \ref{fig:sequential}.
We apply content loss (\ref{eq:cyc}-\ref{eq:ctx}) to the intermediate samples $(y'_m, \bm{b}'_m)$ of current mini-batch $\bm{a}_m$,
as it is just a function of inputs and outputs of the generator $G$.\footnote{
The cycle-consistency loss (\ref{eq:cyc}) needs reconstruction sample $(x''_m, \bm{a}''_m)$.
However, it is just a twice translated current mini-batch sample,
\textit{i.e.,} for the opposite direction generator $G'$, $(x''_m, \bm{a}''_m) = G'(G(x_m, \bm{a}_m))$.}
In contrast, we apply GAN loss (\ref{eq:gan}) to the samples of aggregated mini-batches $(y'_m, \bm{b}'_{1:m})$,
because the network fails to align images and masks when using only a partial subset of instance masks.
We used real/original samples $\{x\}$ with the full set of instance masks only.
Formally, the sequential version of the training loss of InstaGAN is
\begin{equation}
	\mathcal{L}_{\tt InstaGAN-SM} = \sum_{m=1}^M
	\mathcal{L}_{\tt LSGAN}((x, \bm{a}), (y'_m, \bm{b}'_{1:m})) + 
	\mathcal{L}_{\tt content}((x_m, \bm{a}_m), (y'_m, \bm{b}'_m))
	\label{eq:loss-sequential}
\end{equation}
where $\mathcal{L}_{\tt content} = \lambda_{\tt cyc} \mathcal{L}_{\tt cyc}
+ \lambda_{\tt idt} \mathcal{L}_{\tt idt} + \lambda_{\tt ctx} \mathcal{L}_{\tt ctx}$.

We detach every $m$-th iteration of training,
\textit{i.e.,} backpropagating with the mini-batch $\bm{a}_m$,
so that only a fixed GPU memory is required, regardless of the number of training instances.\footnote{
We still recommend users to increase the subset size as long as the GPU memory allows.
This is because too many sequential steps may hurt the permutation-invariance property of our model.}
Hence, the sequential training allows for training with samples containing many instances,
and thus improves the generalization performance.
Furthermore, it 
also improves translation of an image even with a few instances, compared to the one-step approach, due to its data augmentation effect using intermediate samples $(x_m, \bm{a}_m)$.
In our experiments, we divided the instances into mini-batches $\bm{a}_1, \dots, \bm{a}_M$ 
according to the decreasing order of the spatial sizes of instances.
Interestingly, the decreasing order showed a better performance than the random order.
We believe that this is because small instances tend to be occluded by other instances in images, thus often losing their intrinsic shape information. 

\vspace{-0.05in}
\section{Experimental Results}
\vspace{-0.05in}
\label{sec:exp}

\vspace{-0.025in}
\subsection{Image-to-Image Translation Results}
\label{sec:exp-qualitative}
\vspace{-0.025in}

\begin{figure}[t]
    \vspace{-0.2in}
	\centering
	\includegraphics[width=\textwidth]{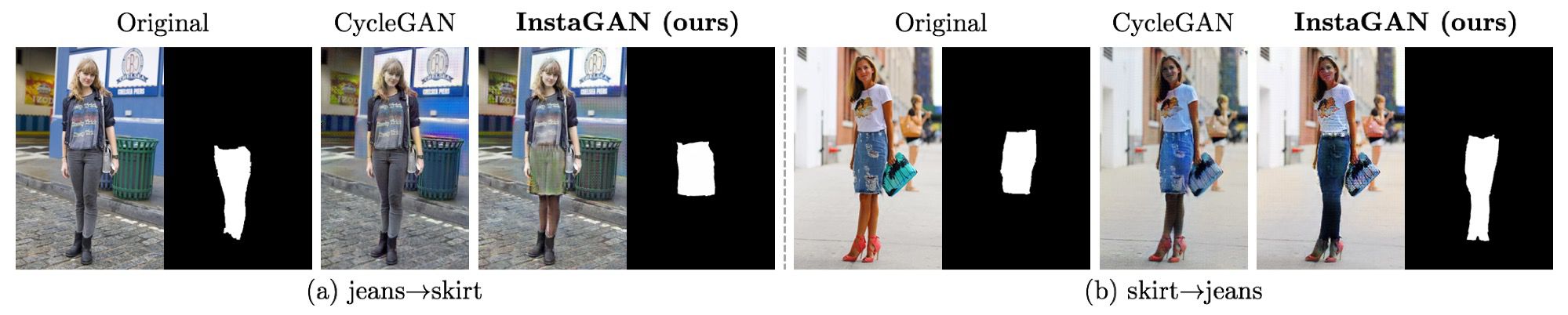}
	\includegraphics[width=\textwidth]{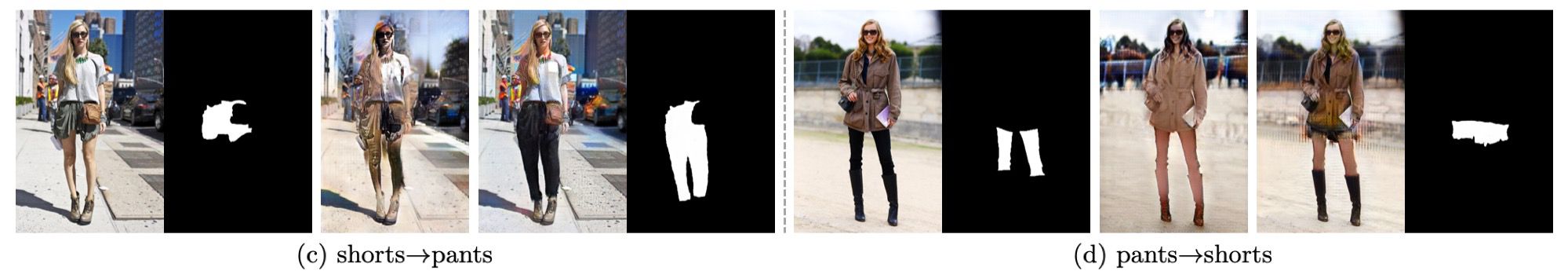}
	\caption{
	Translation results on clothing co-parsing (CCP) \citep{yang2014clothing} dataset.
	} \label{fig:fashion-ccp}
    \vspace{0.05in}
	\includegraphics[width=\textwidth]{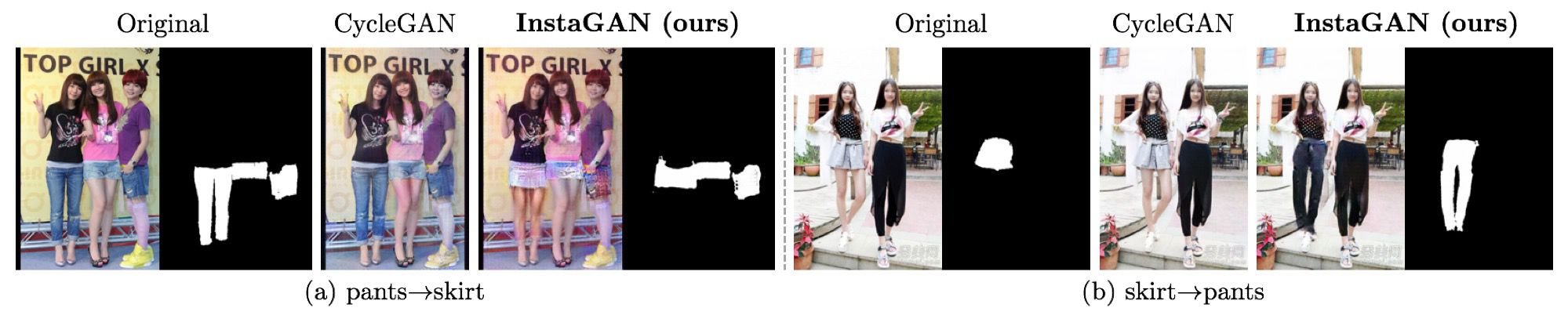}
	\caption{
	Translation results on multi-human parsing (MHP) \citep{zhao2018understanding} dataset.
	} \label{fig:fashion-mhp}	
    \vspace{0.05in}
	\includegraphics[width=\textwidth]{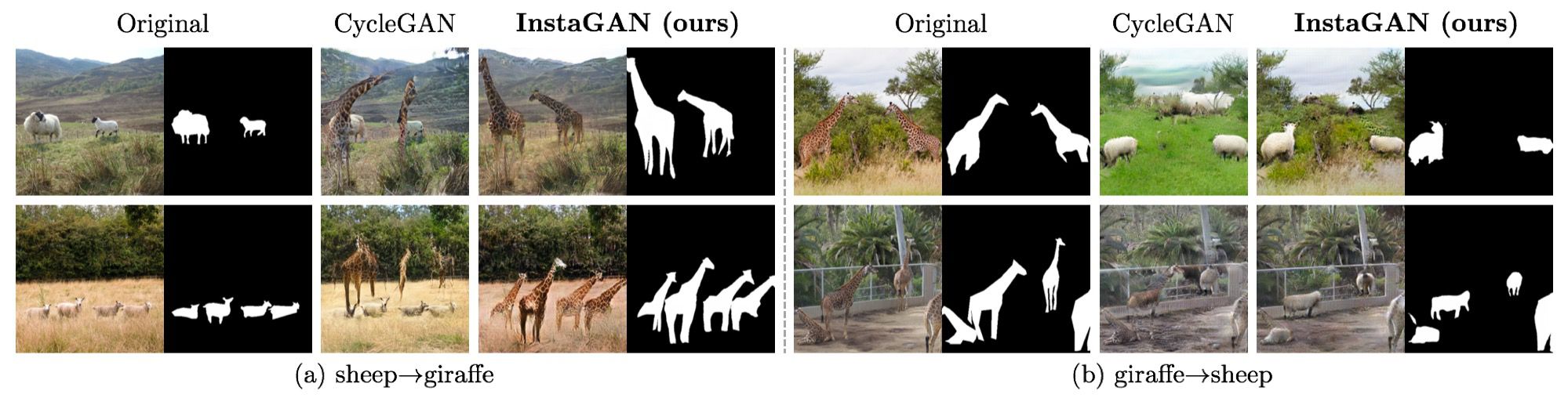}
	\includegraphics[width=\textwidth]{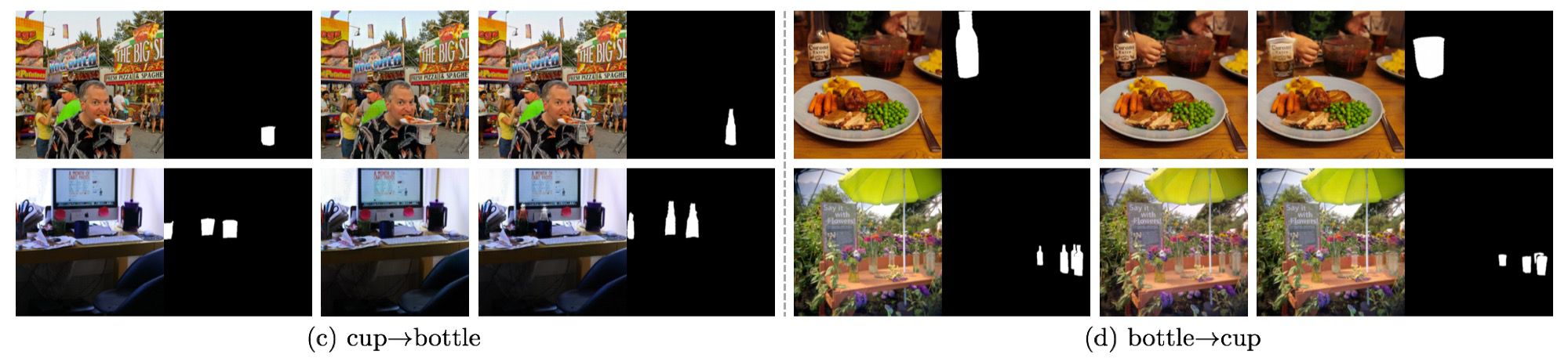}
	\caption{
	Translation results on COCO \citep{lin2014microsoft} dataset.
	} \label{fig:coco}
	\vspace{-0.1in}
\end{figure}

\begin{figure}[t]
    \vspace{-0.2in}
	\centering
	\includegraphics[width=\textwidth]{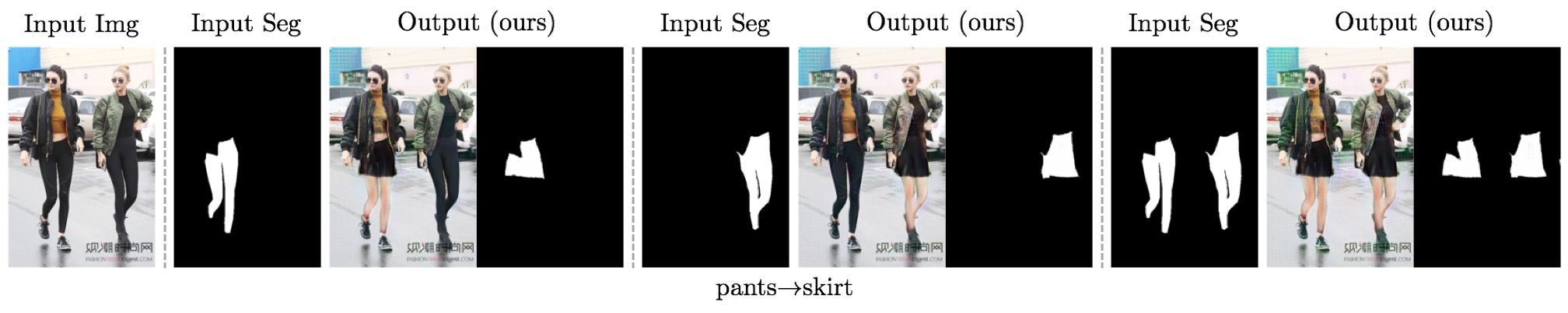}
	\caption{
	Results of InstaGAN varying over different input masks.
	} \label{fig:fashion-mhp-control}
	\vspace{0.05in}
	\includegraphics[width=\textwidth]{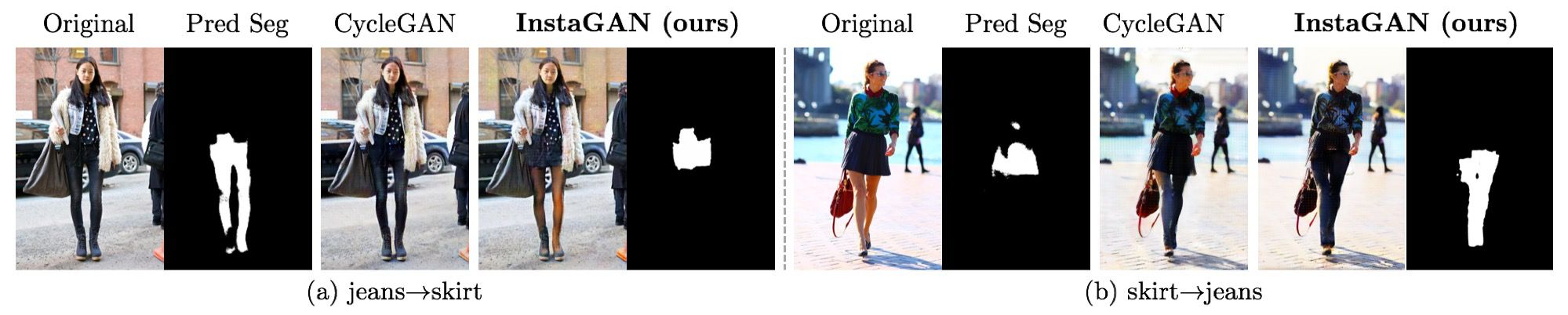}
	\caption{
	Translation results on CCP dataset, using predicted mask for inference.
	} \label{fig:fashion-ccp-pred}
    \vspace{-0.1in}
\end{figure}

We first qualitatively evaluate our method on various datasets.
We compare our model, InstaGAN, with the baseline model, CycleGAN \citep{zhu2017unpaired}.
For fair comparisons, we doubled the number of parameters of CycleGAN, as InstaGAN uses two networks for image and masks, respectively.
We sample two classes from various datasets, including clothing co-parsing (CCP) \citep{yang2014clothing},
multi-human parsing (MHP) \citep{zhao2018understanding}, and MS COCO \citep{lin2014microsoft} datasets,
and use them as the two domains for translation.
In visualizations, we merge all instance masks into one for the sake of compactness.
See Appendix \ref{sec:training-details} for detailed settings for our experiments.
The translation results for three datasets are presented in Figure \ref{fig:fashion-ccp}, \ref{fig:fashion-mhp}, and \ref{fig:coco}, respectively.
While CycleGAN mostly fails, our method generates reasonable shapes of the target instances and keeps the original contexts by focusing on the
instances via the context preserving loss.
For example, see the results on sheep$\leftrightarrow$giraffe in Figure \ref{fig:coco}.
CycleGAN often generates sheep-like instances but loses the original background.
InstaGAN  not only generates better sheep or giraffes, but also preserves the layout of the original instances,
\textit{i.e.,} the looking direction (left, right, front) of sheep and giraffes are consistent after translation.
More experimental results are presented in Appendix \ref{sec:more-results}.
Code and results are available in \url{https://github.com/sangwoomo/instagan}.

On the other hand, our method can control the instances to translate by conditioning the input, as shown in Figure \ref{fig:fashion-mhp-control}.
Such a control is impossible under
CycleGAN. 
We also note that we focus on complex (multi-instance transfiguration) tasks to emphasize the advantages of our method.
Nevertheless, our method is also attractive to use even for simple tasks (\textit{e.g.,} horse$\leftrightarrow$zebra)
as it reduces false positives/negatives via the context preserving loss and enables to control translation.
We finally emphasize that our method showed good results even when we use predicted segmentation for inference,
as shown in Figure \ref{fig:fashion-ccp-pred}, and this can reduce the cost of collecting mask labels in practice.\footnote{
For the results in Figure \ref{fig:fashion-ccp-pred}, we trained a pix2pix \citep{isola2017image} model to predict a single mask from an image,
but one can also utilize recent methods to predict instance masks
in supervised \citep{he2017mask} or weakly-supervised \citep{zhou2018weakly} way.}


Finally, we also quantitatively evaluate the translation performance of our method.
We measure the classification score, the ratio of images predicted as the target class by a pretrained classifier.
Specifically, we fine-tune the final layers of the
ImageNet \citep{deng2009imagenet} pretrained VGG-16 \citep{simonyan2014very} network,
as a binary classifier for each domain.
Table \ref{tab:cls-ccp} and Table \ref{tab:cls-coco} in Appendix \ref{sec:cls-score} show the classification scores
for CCP and COCO datasets, respectively.
Our method outperforms CycleGAN in all classification experiments,
\textit{e.g.,} ours achieves 23.2\% accuracy for the pants$\to$shorts task,
while CycleGAN obtains only 8.5\%.


\vspace{-0.025in}
\subsection{Ablation Study}
\label{sec:exp-ablation}
\vspace{-0.025in}

\begin{figure}[t]
	\centering
	\includegraphics[width=\textwidth]{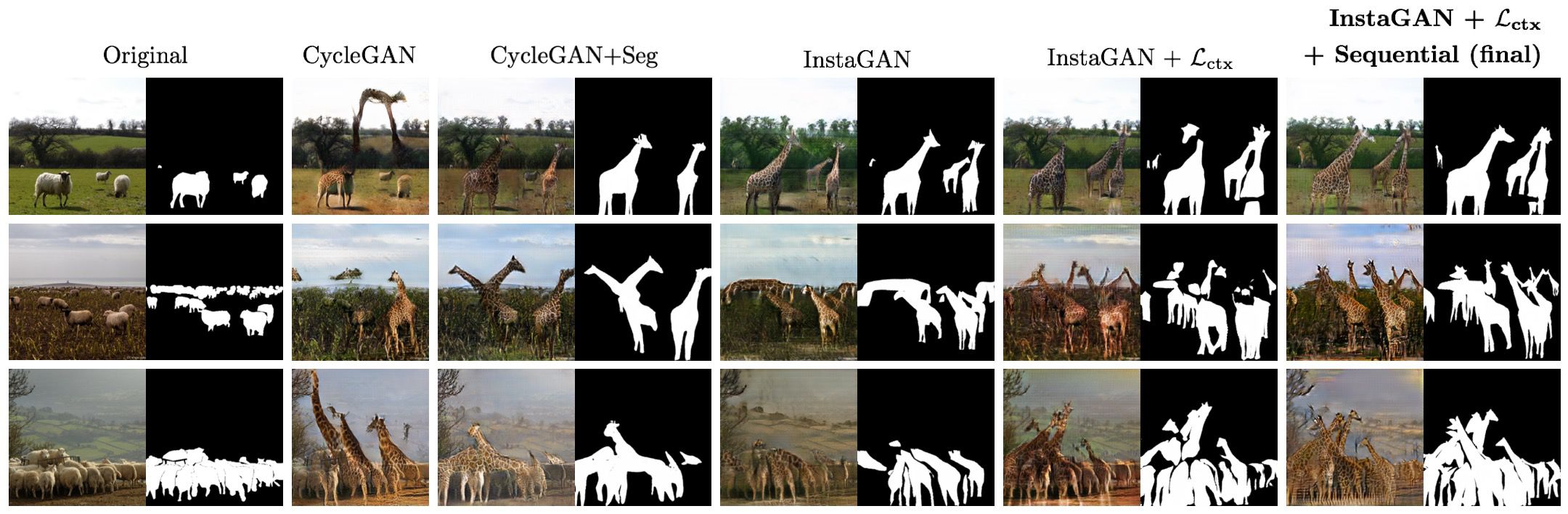}
	\caption{
	Ablation study on the effect of each component of our method: the InstaGAN architecture, 
	the context preserving loss, and the sequential mini-batch inference/training algorithm,
	which are denoted as InstaGAN, $\mathcal{L}_\text{ctx}$, and Sequential, respectively.
	} \label{fig:ablation-1}
\end{figure}	

\begin{figure}[t]
    \vspace{-0.2in}
	\includegraphics[width=\textwidth]{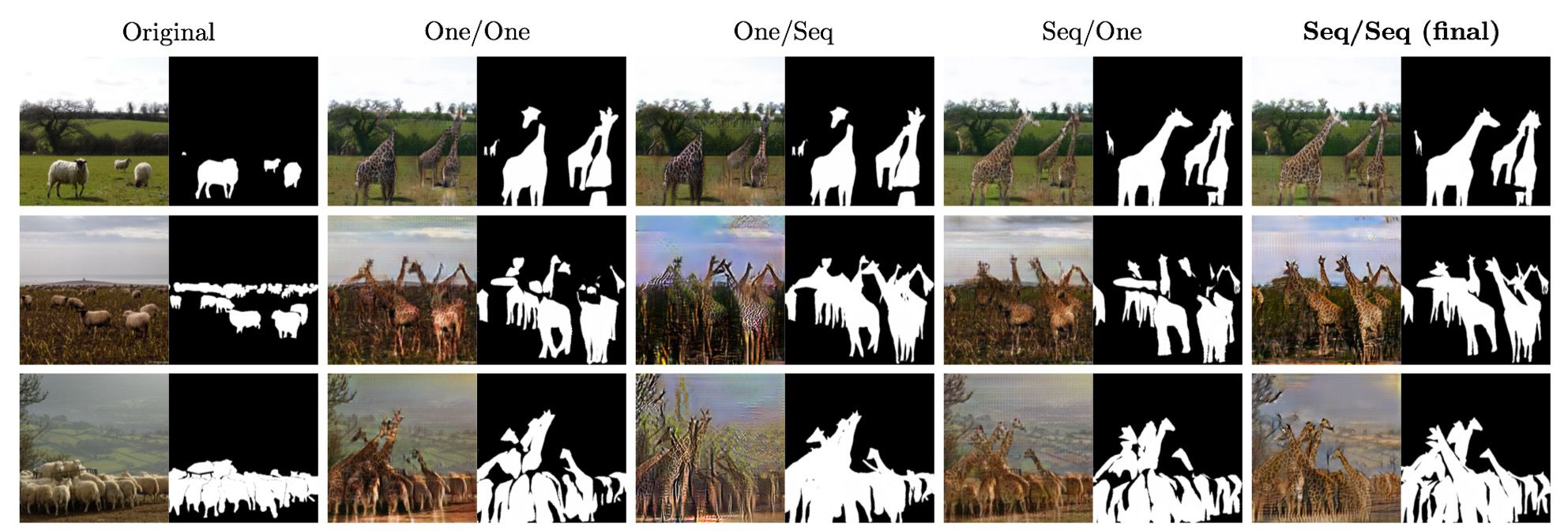}
	\caption{
	Ablation study on the effects of the sequential mini-batch inference/training technique.
	The left and right side of title indicates which method used for training and inference, respectively, where
	``One'' and ``Seq'' indicate the one-step and sequential schemes, respectively.
	} \label{fig:ablation-2}
	\vspace{-0.1in}
\end{figure}

We now investigate the effects of each component of our proposed method in Figure \ref{fig:ablation-1}.
Our method is composed of the InstaGAN architecture, 
the context preserving loss $\mathcal{L}_\text{ctx}$,
and the sequential mini-batch inference/training technique.
We progressively add each component to the baseline model, CycleGAN (with doubled parameters).
First, we study the effect of our architecture.
For fair comparison, we train a CycleGAN model with an additional input channel,
which translates the mask-augmented image, hence we call it CycleGAN+Seg.
Unlike our architecture which translates the set of instance masks,
CycleGAN+Seg translates the union of all masks at once.
Due to this, CycleGAN+Seg fails to translate some instances and often merge them.
On the other hand, our architecture keeps every instance and disentangles better.
Second, we study the effect of the context preserving loss: it not only preserves the background better (row 2),
but also improves the translation results as it regularizes the mapping (row 3).
Third, we study the effect of our sequential translation:
it not only improves the generalization performance (row 2,3)
but also improves the translation results on few instances, via data augmentation (row 1).

Finally, Figure \ref{fig:ablation-2} reports how much the sequential translation,
denoted by ``Seq'', is effective in inference and training, compared to the one-step approach, denoted by ``One''.
For the one-step training,
we consider only two instances, as it is the maximum number affordable for our machines.
On the other hand, for the sequential training,  
we sequentially train two instances twice, \textit{i.e.,} images of four instances.
For the one-step inference, we translate the entire set at once,
and for the sequential inference, we sequentially translate two instances at each iteration.
We find that our sequential algorithm is effective for both training and inference:
(a) training/inference = One/Seq shows blurry results as intermediate data have not shown during training and stacks noise as the iteration goes,
and (b) Seq/One shows poor generalization performance for multiple instances as
the one-step inference for many instances is not shown in training (due to a limited GPU memory).


\vspace{-0.05in}
\section{Conclusion}
\vspace{-0.05in}

We have proposed a novel method incorporating the set of instance attributes for image-to-image translation.
The experiments on different datasets have shown successful image-to-image translation on the challenging tasks of multi-instance transfiguration,
including new tasks, \textit{e.g.,} translating jeans to skirt in fashion images.
We remark that our ideas utilizing the set-structured side information  
have potential to be applied to other cross-domain generations tasks,
\textit{e.g.,} neural machine translation or video generation.
Investigating new tasks and new information could be an interesting research direction in the future.

\subsubsection*{Acknowledgments}
This work was supported by the National Research Council of Science \& Technology (NST) grant by the Korea government (MSIP) (No. CRC-15-05-ETRI),
by the ICT R\&D program of MSIT/IITP [2016-0-00563, Research on Adaptive Machine Learning Technology Development for Intelligent Autonomous Digital Companion],
and also by Basic Science Research Program (NRF-2017R1E1A1A01077999) through the National Research Foundation of Korea (NRF) funded by the Ministry of Science, ICT.

\bibliography{main}
\bibliographystyle{iclr2019_conference}

\newpage
\appendix
\section{Architecture Details}
\label{sec:architecture-details}
We adopted the network architectures of CycleGAN \citep{zhu2017unpaired} as the building blocks for our proposed model.
In specific, we adopted ResNet $9$-blocks generator \citep{johnson2016perceptual, he2016deep} and PatchGAN \citep{isola2017image} discriminator.
ResNet generator is composed of downsampling blocks, residual blocks, and upsampling blocks.
We used downsampling blocks and residual blocks for encoders, and used upsampling blocks for generators.
On the other hand, PatchGAN discriminator is composed of 5 convolutional layers, including normalization and non-linearity layers.
We used the first 3 convolution layers for feature extractors, and the last 2 convolution layers for classifier.
We preprocessed instance segmentation as a binary foreground/background mask, hence simply used it as an 1-channel binary image.
Also, since we concatenated two or three features to generate the final outputs, we doubled or tripled the input dimension of those architectures.
Similar to prior works \citep{johnson2016perceptual, zhu2017unpaired},
we applied Instance Normalization (IN) \citep{ulyanov2016instance} for both generators and discriminators.
In addition, we observed that applying Spectral Normalization (SN) \citep{miyato2018spectral}
for discriminators significantly improves the performance, although we used LSGAN \citep{mao2017least},
while the original motivation of SN was to enforce Lipschitz condition to match with the theory of WGAN \citep{arjovsky2017wasserstein, gulrajani2017improved}.
We also applied SN for generators as suggested in Self-Attention GAN \citep{zhang2018self}, but did not observed gain for our setting.

\section{Training Details}
\label{sec:training-details}

For all the experiments, we simply set $\lambda_\text{cyc} = 10$, $\lambda_\text{idt} = 10$, and $\lambda_\text{ctx} = 10$ for our loss (\ref{eq:loss}).
We used Adam \citep{kingma2014adam} optimizer with batch size 4, training with 4 GPUs in parallel.
All networks were trained from scratch, with learning rate of 0.0002 for $G$ and 0.0001 for $D$, and $\beta_1 = 0.5$, $\beta_2 = 0.999$ for the optimizer.
Similar to CycleGAN \citep{zhu2017unpaired}, we kept learning rate for first 100 epochs and linearly decayed to zero for next 100 epochs
for multi-human parsing (MHP) \citep{zhao2018understanding} and COCO \citep{lin2014microsoft} dataset,
and kept learning rate for first 400 epochs and linearly decayed for next 200 epochs
for clothing co-parsing (CCP) \citep{yang2014clothing} dataset, as it contains smaller number of samples.
We sampled two classes from the datasets above, and used it as two domains for translation.
We resized images with size 300$\times$200 (height$\times$width) for CCP dataset, 240$\times$160 for MHP dataset, and 200$\times$200 for COCO dataset, respectively.

\section{Trend of Translation Results}
\label{sec:epoch-trends}

\begin{figure}[H]
	\centering
	\includegraphics[width=\textwidth]{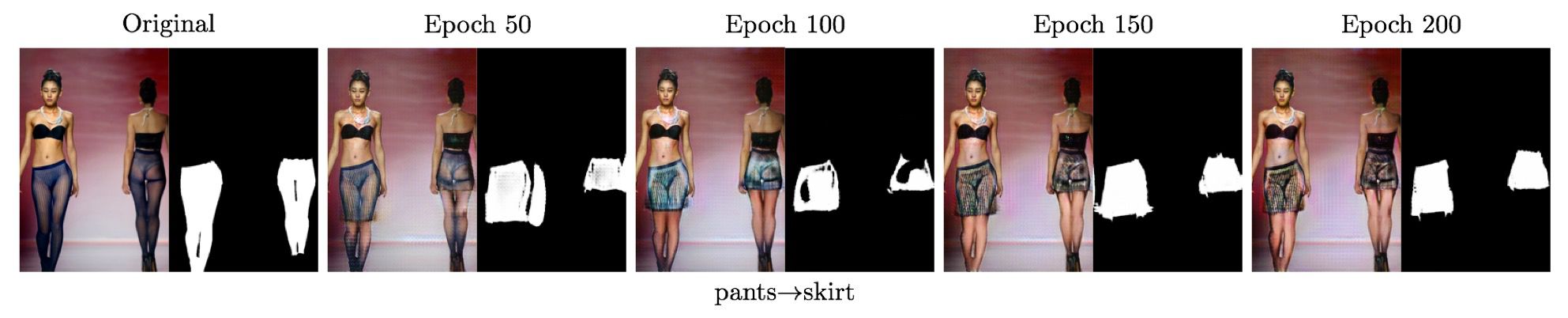}
	\caption{
	Trend of the translation results of our method over epoch increases.
	} \label{fig:fashion-mhp-trend}
\end{figure}

We tracked the trend of translation results over epoch increases, as shown in Figure \ref{fig:fashion-mhp-trend}.
Both image and mask smoothly adopted to the target instances.
For example, the remaining parts in legs slowly disappears,
and the skirt slowly constructs the triangular shapes.

\newpage
\section{Quantitative Results}
\label{sec:cls-score}


We evaluated the classification score for CCP and COCO dataset.
Unlike CCP dataset, COCO dataset suffers from the false positive problem,
that the classifier fails to determine if the generator produced target instances on the right place. 
To overcome this issue, we measured the masked classification score,
where the input images are masked by the corresponding segmentations.
We note that CycleGAN and our method showed comparable results for the na\"ive classification score,
but ours outperformed for the masked classification score, as it reduces the false positive problem.

\begin{table}[H]
\caption{
Classification score for CCP dataset.
} \label{tab:cls-ccp}
\centering
\begin{tabular}{lcccc|cccc}
	\toprule
	& \multicolumn{2}{c}{jeans$\to$skirt} & \multicolumn{2}{c}{skirt$\to$jeans} & \multicolumn{2}{|c}{shorts$\to$pants} & \multicolumn{2}{c}{pants$\to$shorts} \\
	\cmidrule{2-5} \cmidrule{6-9}
	& train & test & train & test & train & test & train & test \\
	\midrule
	Real & 0.970 & 0.888 & 0.982 & 0.946 & 1.000 & 0.984 & 0.990 & 0.720 \\
	\midrule
	CycleGAN & 0.465 & 0.371 & 0.561 & 0.483 & 0.845 & 0.524 & 0.305 & 0.085 \\
	\textbf{InstaGAN (ours)} & \textbf{0.665} & \textbf{0.600} & \textbf{0.658} & \textbf{0.540} & \textbf{0.898} & \textbf{0.768} & \textbf{0.373} & \textbf{0.232} \\
	\bottomrule
\end{tabular}
\vspace{0.1in}
\caption{
Classification score (masked) for COCO dataset.
} \label{tab:cls-coco}
\begin{tabular}{lcccc|cccc}
	\toprule
    & \multicolumn{2}{c}{sheep$\to$giraffe} & \multicolumn{2}{c}{giraffe$\to$sheep} & \multicolumn{2}{|c}{cup$\to$bottle} & \multicolumn{2}{c}{bottle$\to$cup} \\
	\cmidrule{2-5} \cmidrule{6-9}
	& train & test & train & test & train & test & train & test \\
	\midrule
	Real & 0.891 & 0.911 & 0.925 & 0.930 & 0.746 & 0.723 & 0.622 & 0.566 \\
	\midrule
	CycleGAN & 0.313 & 0.594 & 0.291 & 0.512 & 0.368 & 0.403 & 0.290 & 0.275 \\
	\textbf{InstaGAN (ours)} & \textbf{0.406} & \textbf{0.781} & \textbf{0.355} & \textbf{0.642} & \textbf{0.443} & \textbf{0.465} & \textbf{0.322} & \textbf{0.333} \\
	\bottomrule
\end{tabular}
\end{table}

\section{More Translation Results}
\label{sec:more-results}

We present more qualitative results in high resolution images.

\begin{figure}[H]
	\centering
    \includegraphics[width=\textwidth]{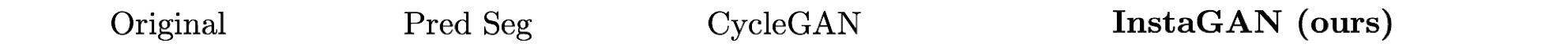}
	\includegraphics[width=\textwidth]{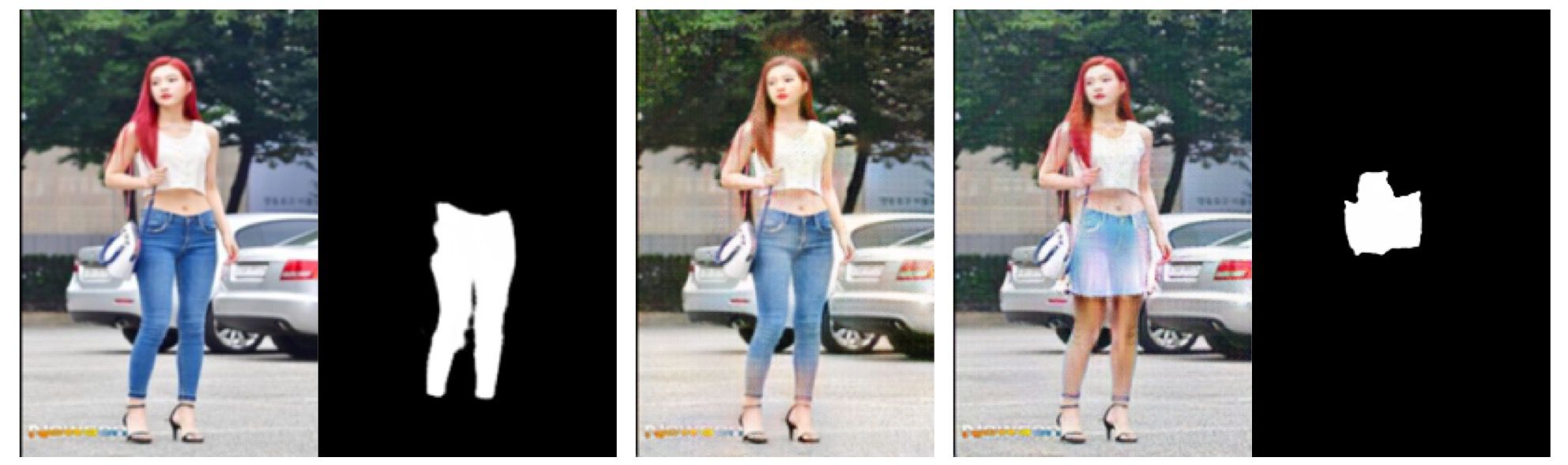}
	\includegraphics[width=\textwidth]{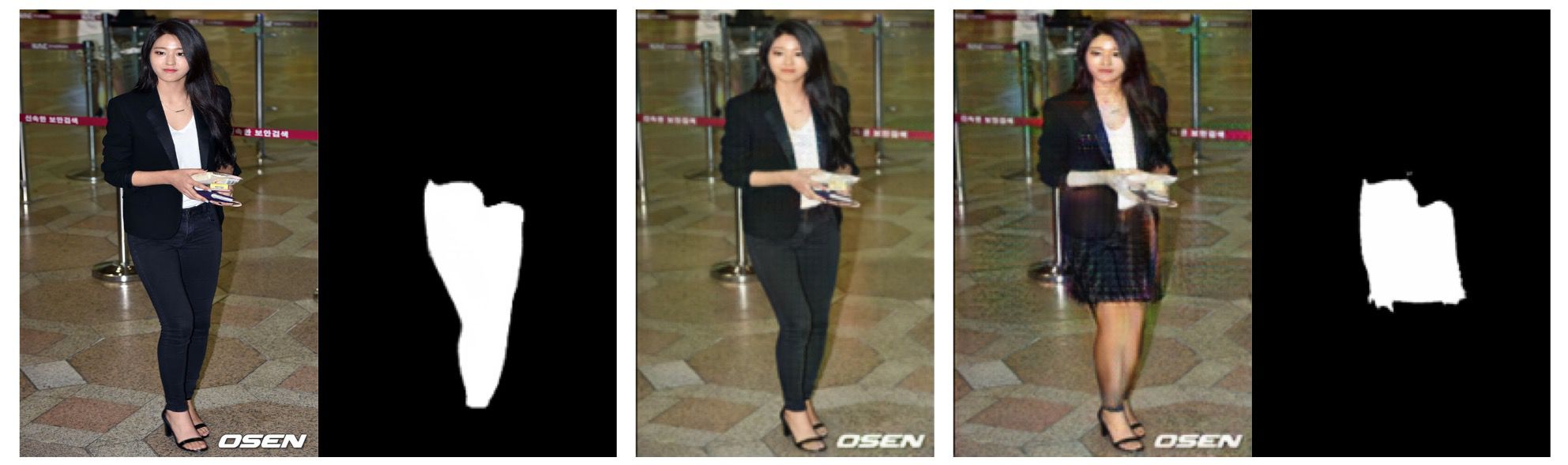}
	\caption{
	Translation results for images searched from Google to test the generalization performance of our model.
	We used a pix2pix \citep{isola2017image} model to predict the segmentation.
	} \label{fig:kpop}
\end{figure}

\begin{figure}[H]
	\centering
	\includegraphics[width=\textwidth]{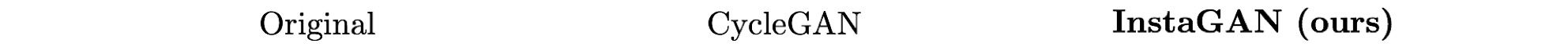}
	\includegraphics[width=\textwidth]{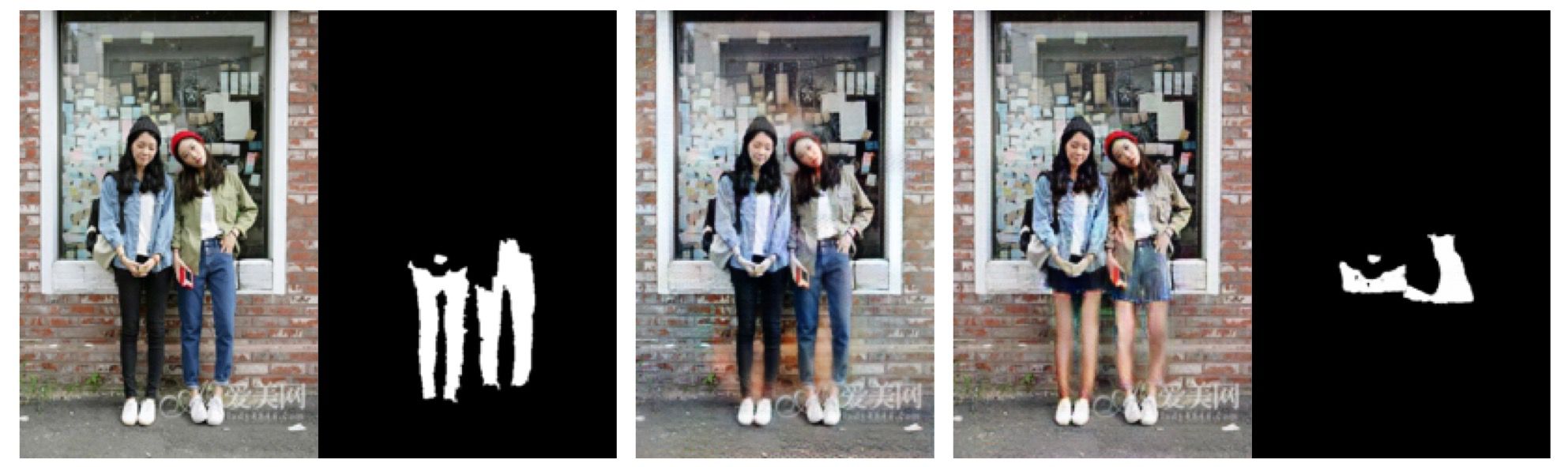}
	\includegraphics[width=\textwidth]{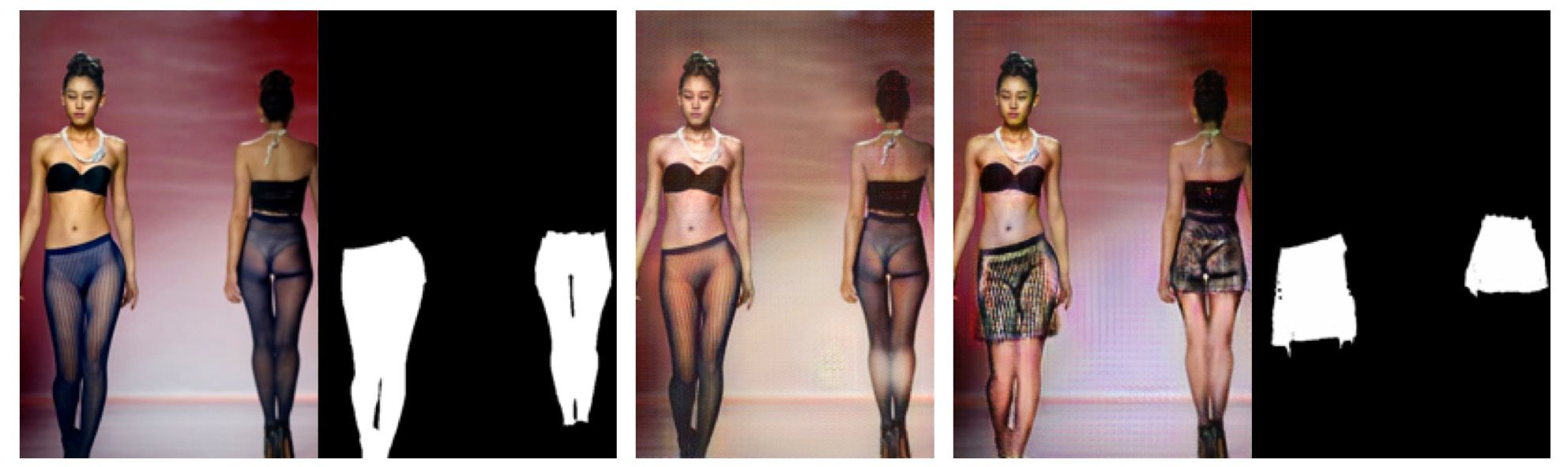}
	\includegraphics[width=\textwidth]{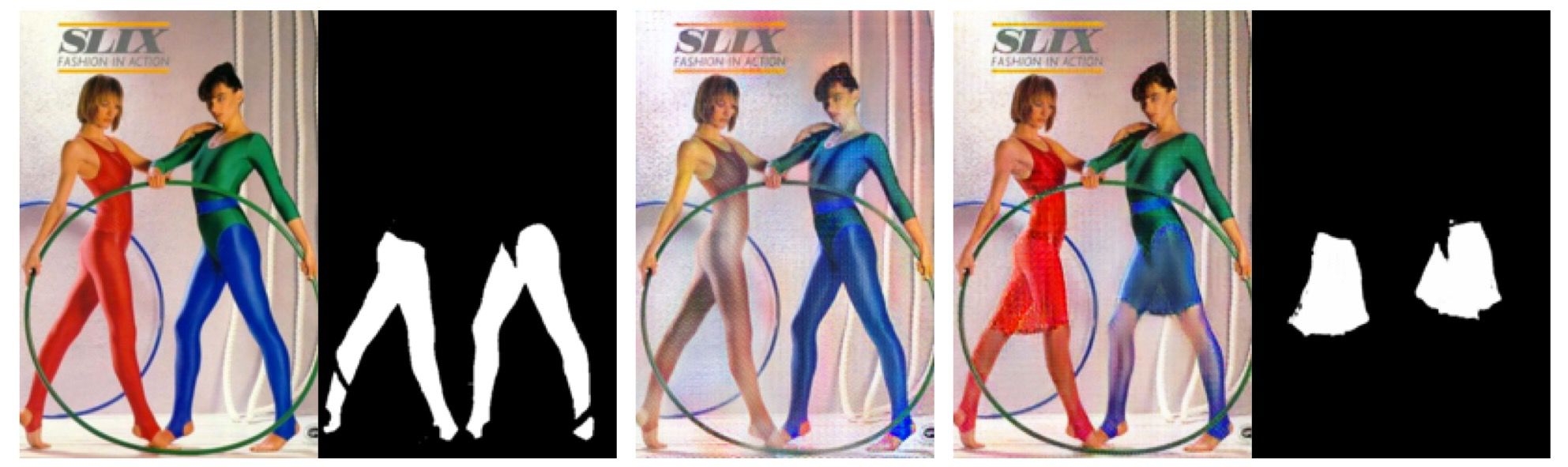}
	\includegraphics[width=\textwidth]{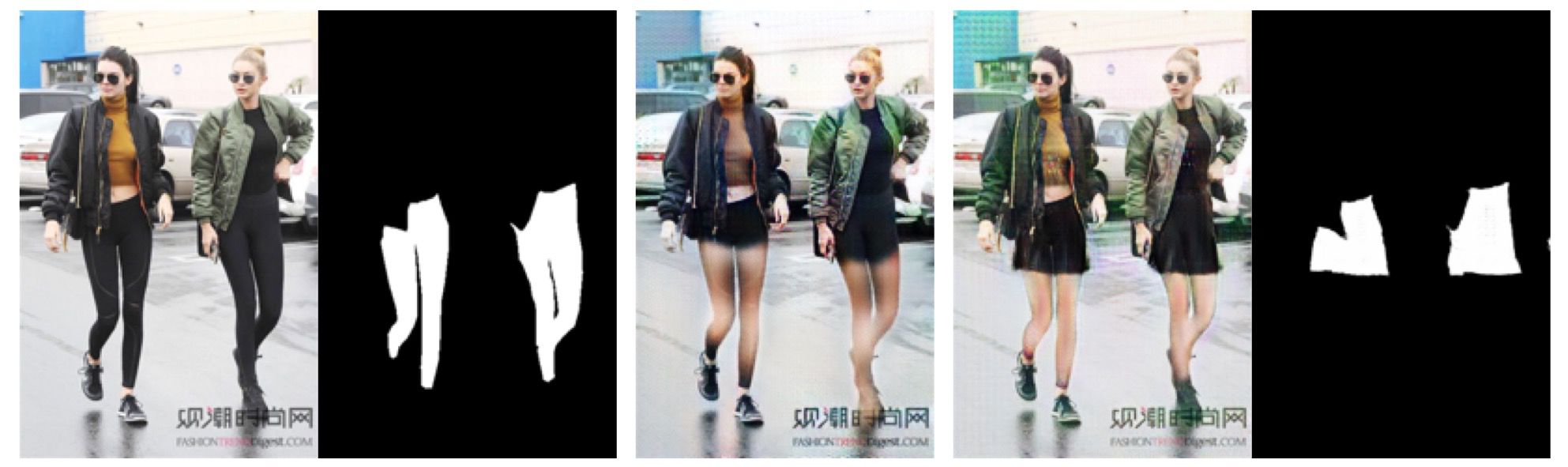}
	\includegraphics[width=\textwidth]{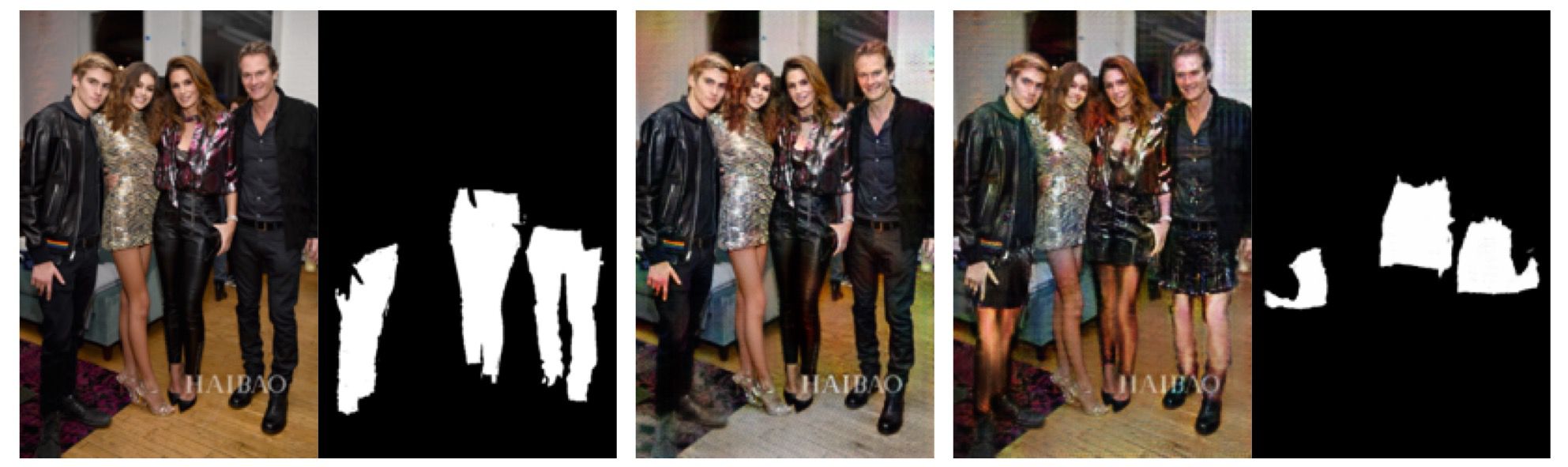}
	\caption{
	More translation results on MHP dataset (pants$\to$skirt).
	} \label{fig:more-mhp-1}
\end{figure}

\begin{figure}[H]
	\centering
	\includegraphics[width=\textwidth]{figure/appendix-label}
	\includegraphics[width=\textwidth]{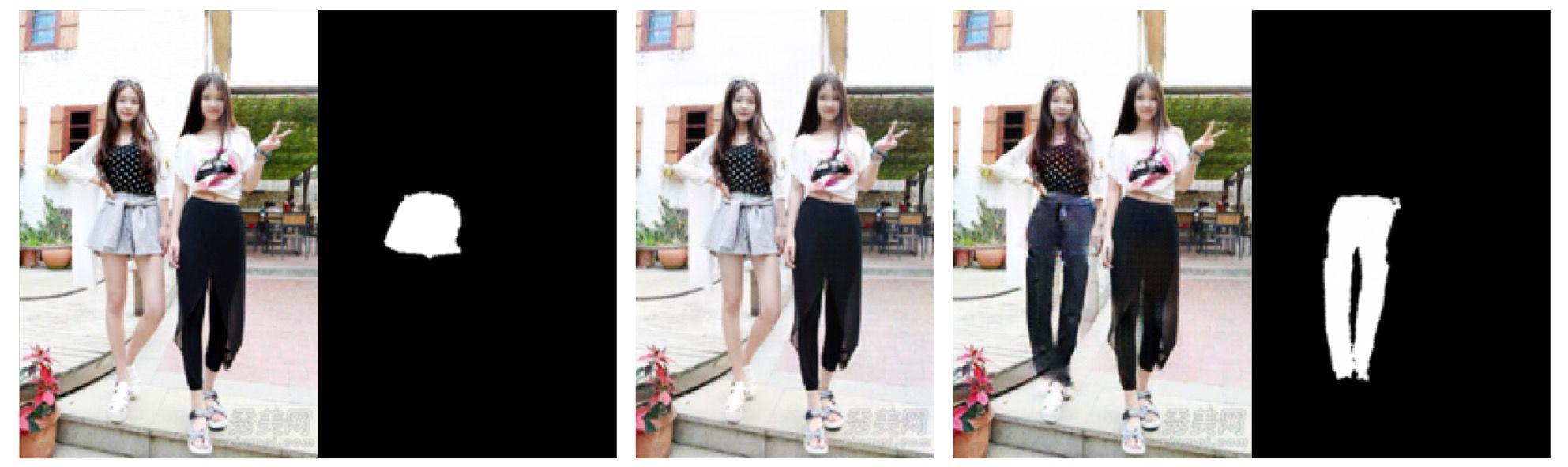}
	\includegraphics[width=\textwidth]{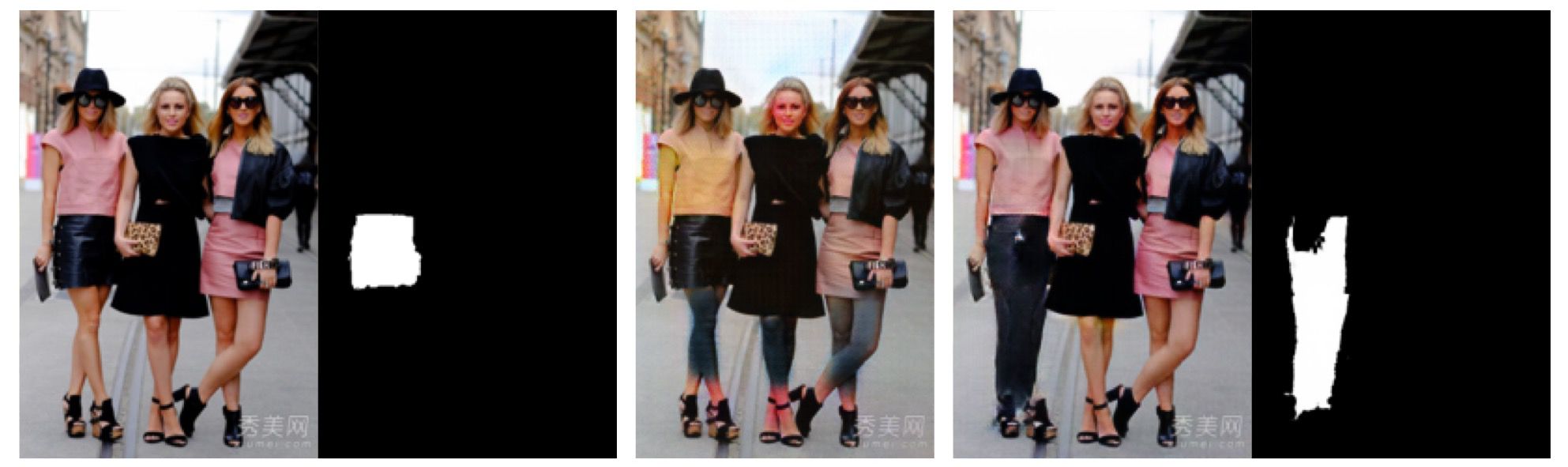}
	\includegraphics[width=\textwidth]{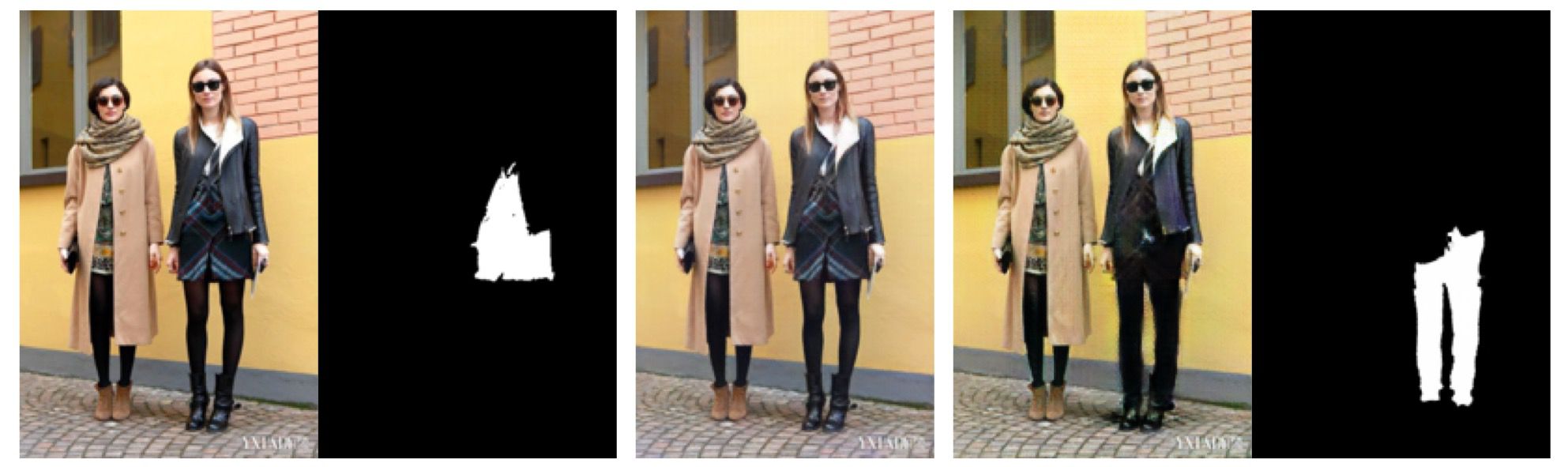}
	\includegraphics[width=\textwidth]{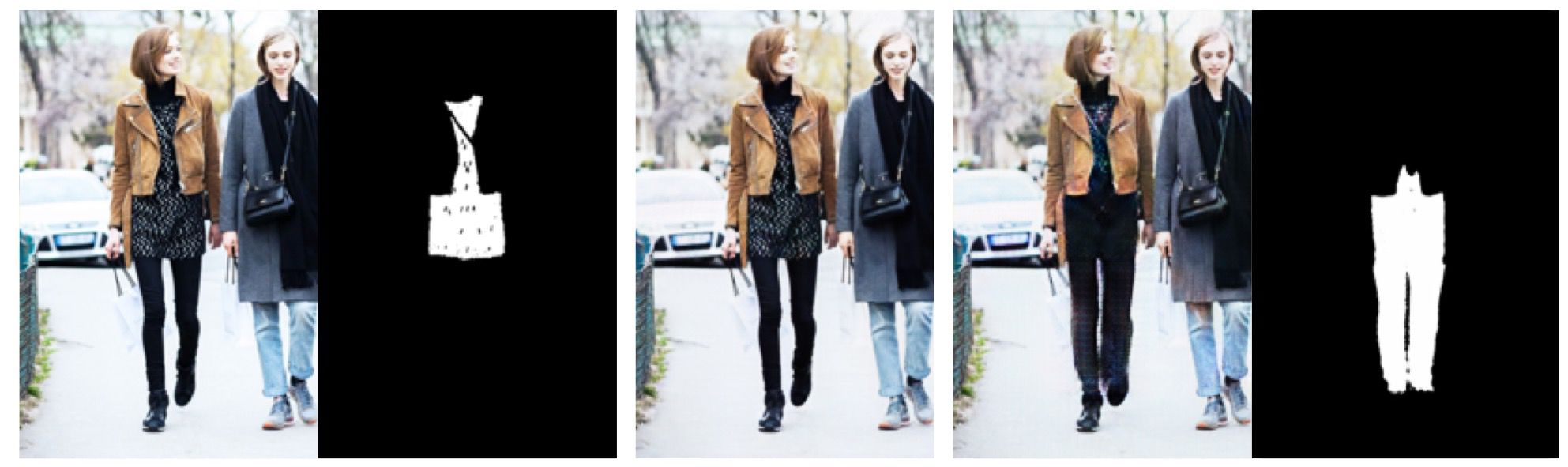}
	\includegraphics[width=\textwidth]{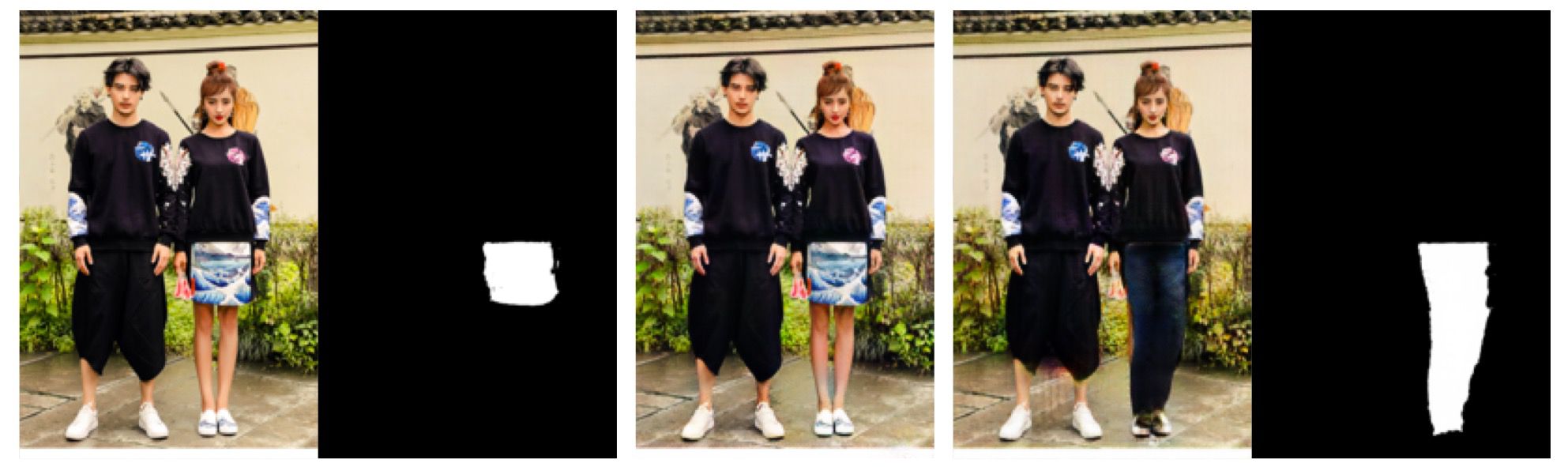}
	\caption{
	More translation results on MHP dataset (skirt$\to$pants).
	} \label{fig:more-mhp-2}
\end{figure}

\begin{figure}[H]
	\centering
	\includegraphics[width=\textwidth]{figure/appendix-label}
	\includegraphics[width=\textwidth]{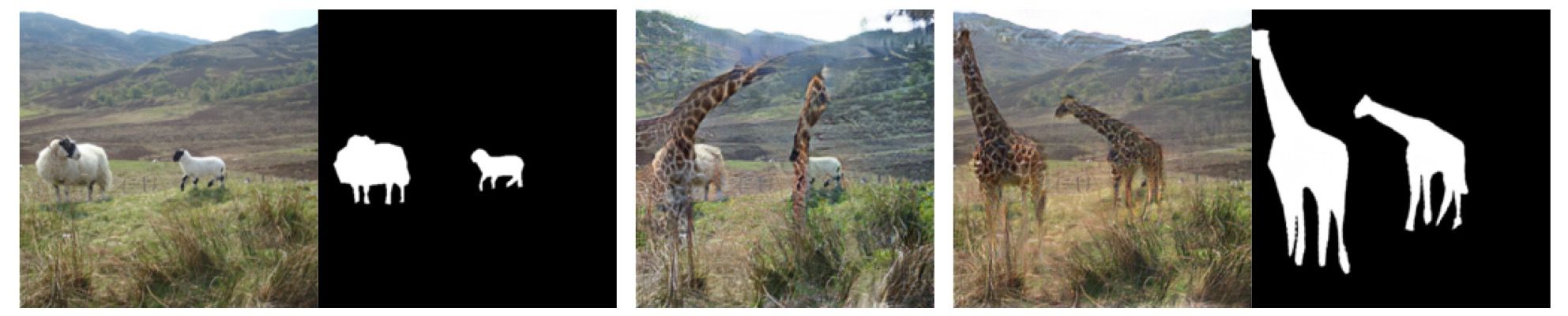}
    \includegraphics[width=\textwidth]{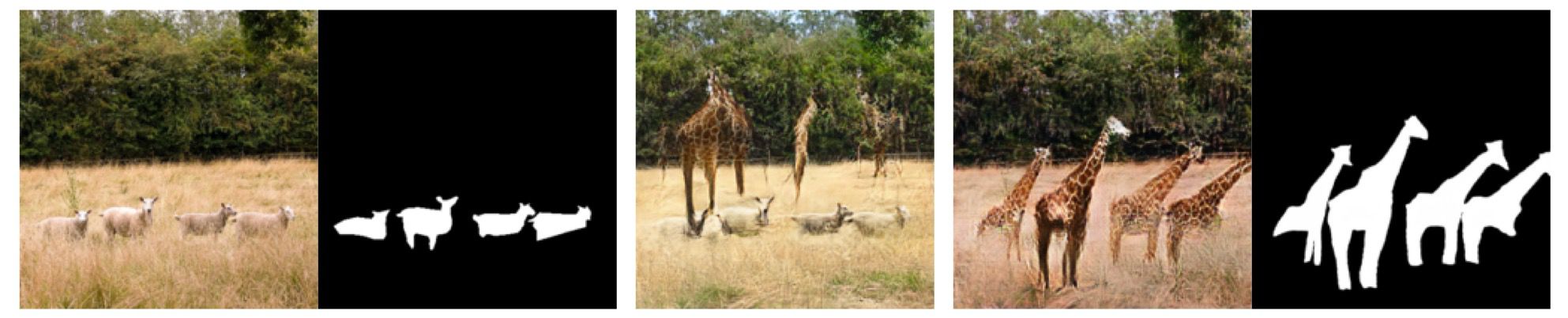}
    \includegraphics[width=\textwidth]{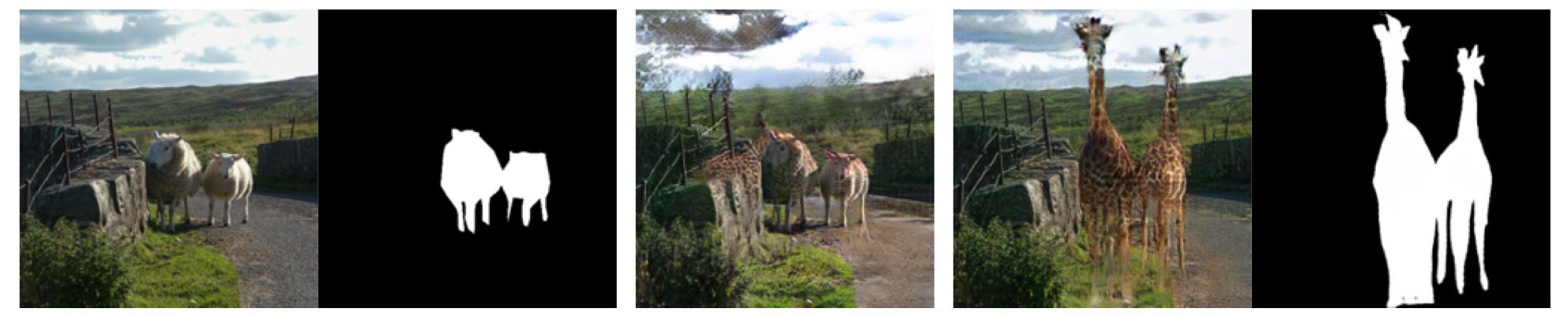}
    \includegraphics[width=\textwidth]{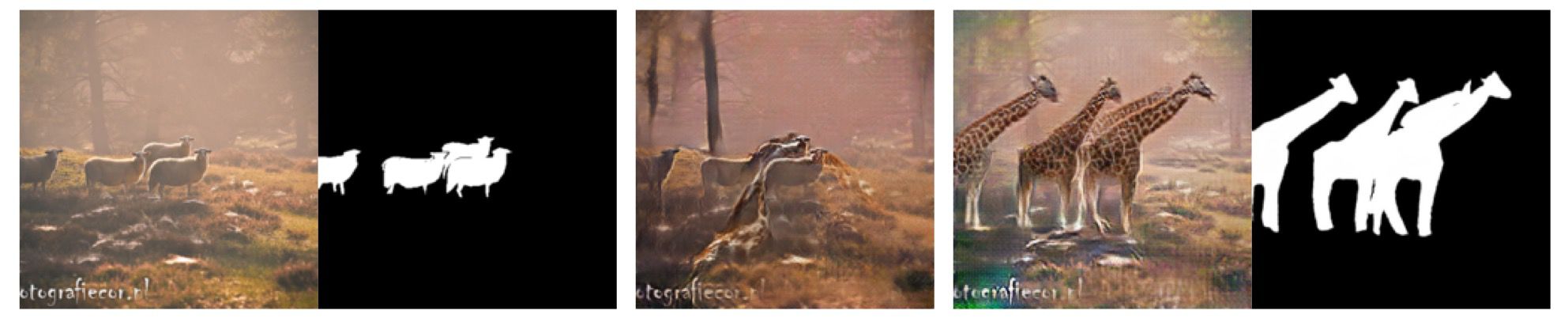}
    \includegraphics[width=\textwidth]{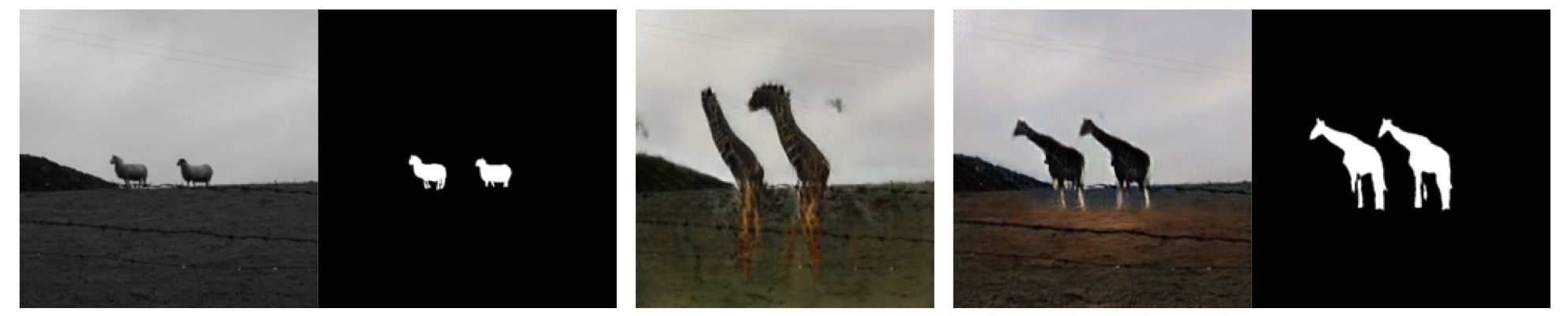}
    \includegraphics[width=\textwidth]{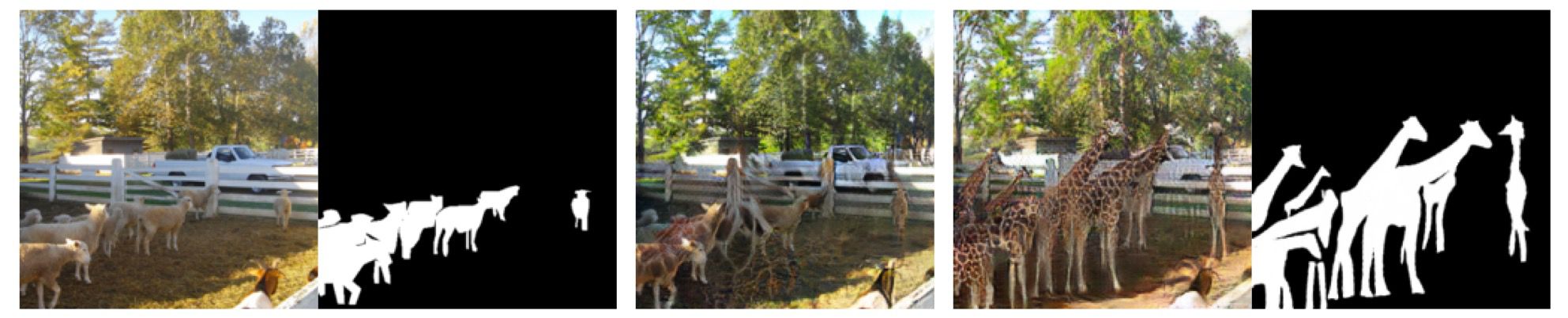}
    \includegraphics[width=\textwidth]{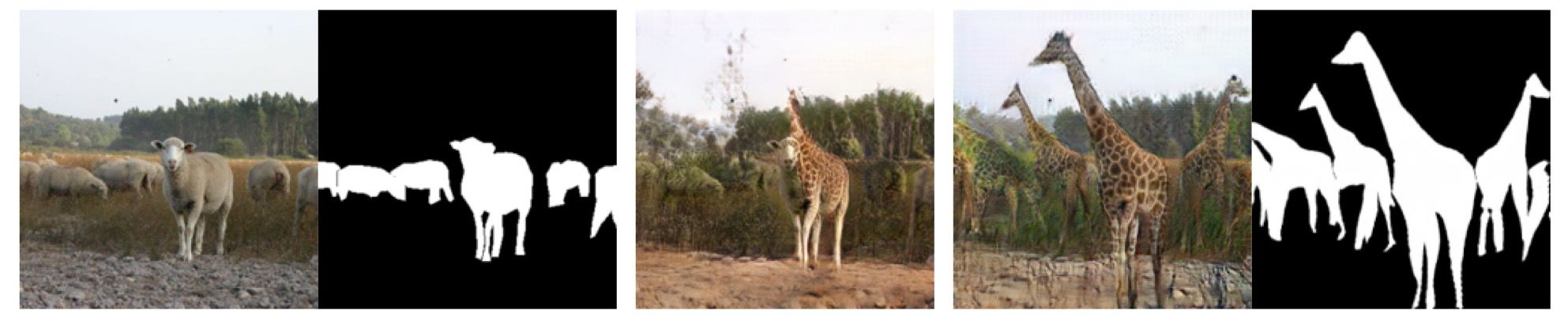}
	\caption{
	More translation results on COCO dataset (sheep$\to$giraffe).
	} \label{fig:more-coco-1}
\end{figure}

\begin{figure}[H]
	\centering
	\includegraphics[width=\textwidth]{figure/appendix-label}
	\includegraphics[width=\textwidth]{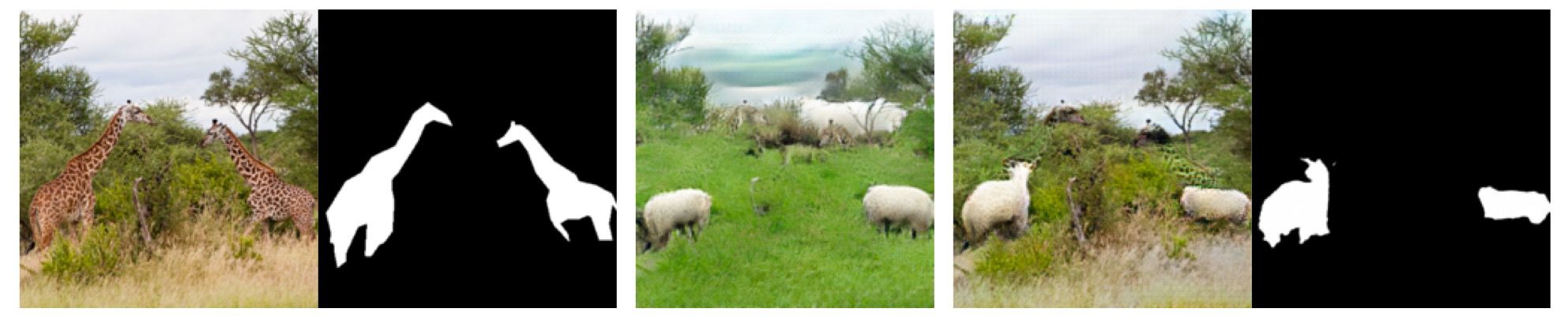}
    \includegraphics[width=\textwidth]{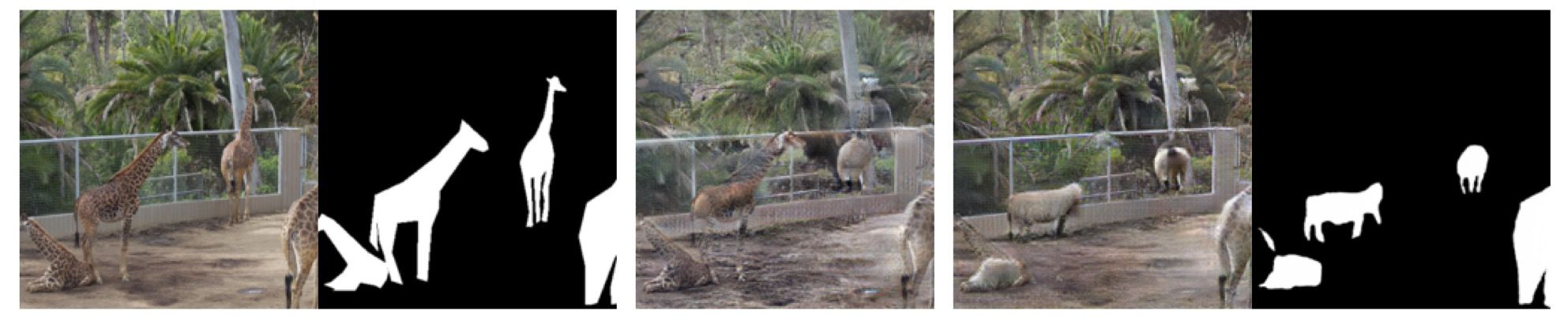}
    \includegraphics[width=\textwidth]{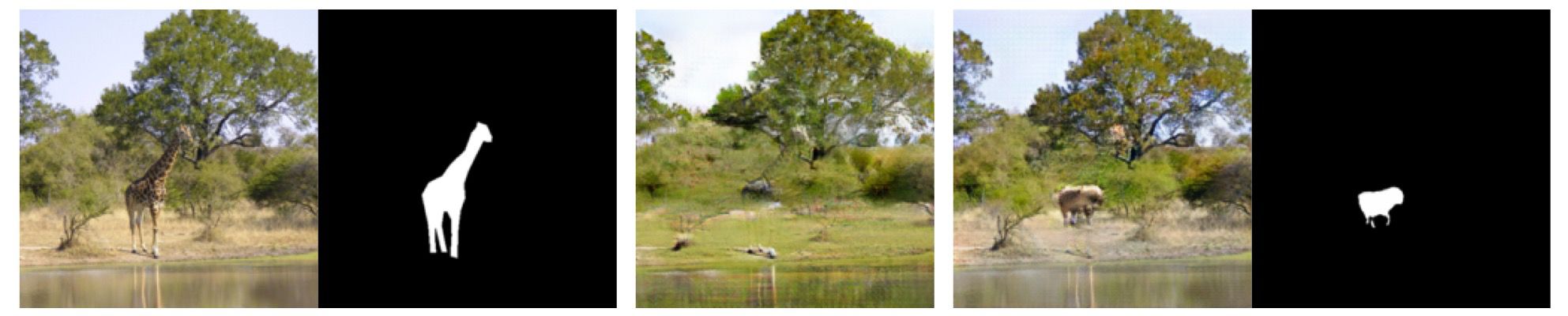}
    \includegraphics[width=\textwidth]{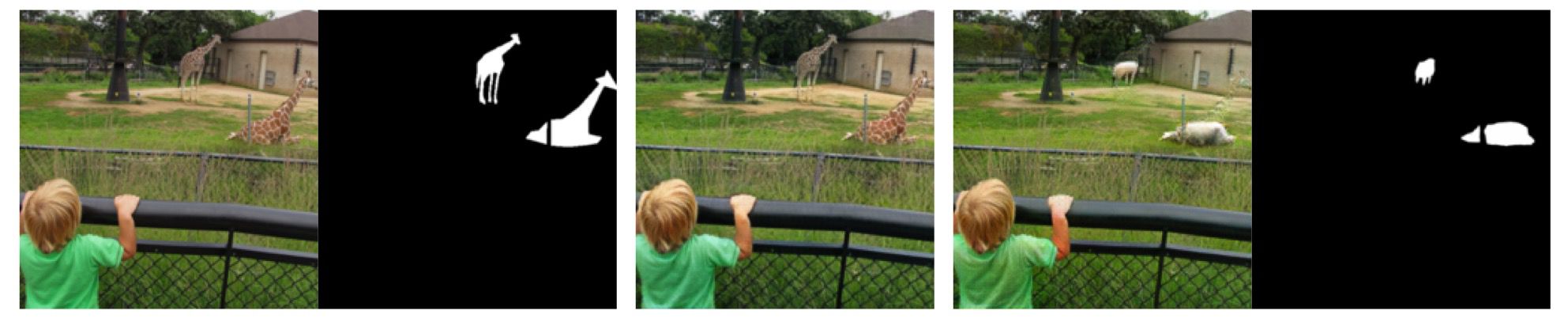}
    \includegraphics[width=\textwidth]{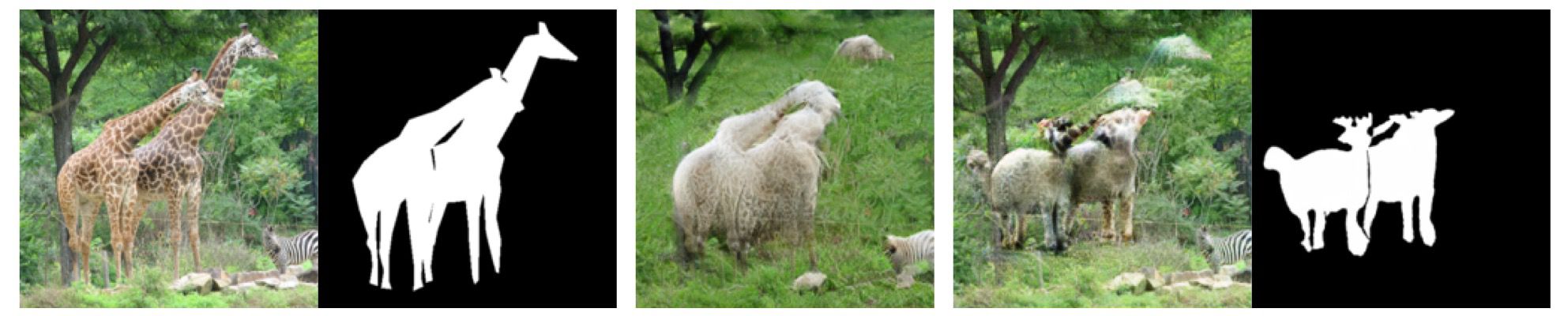}
    \includegraphics[width=\textwidth]{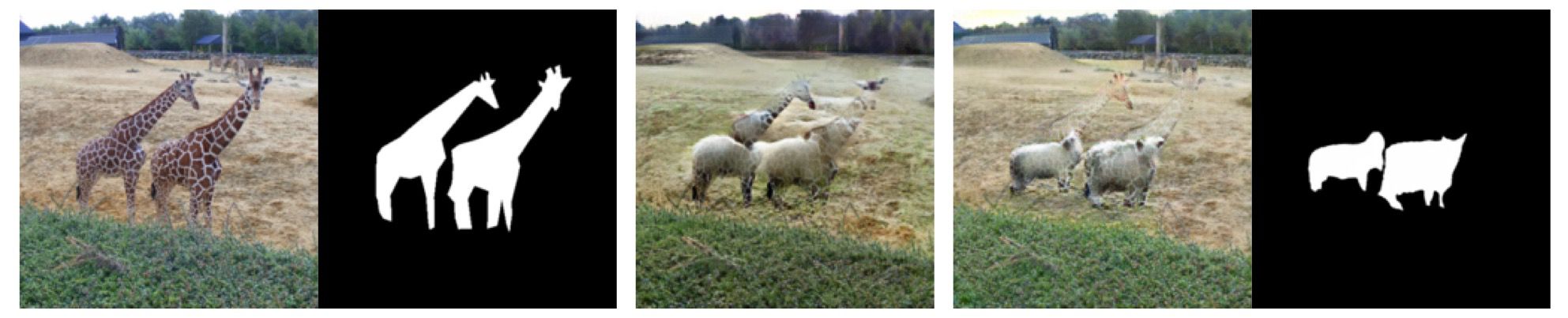}
    \includegraphics[width=\textwidth]{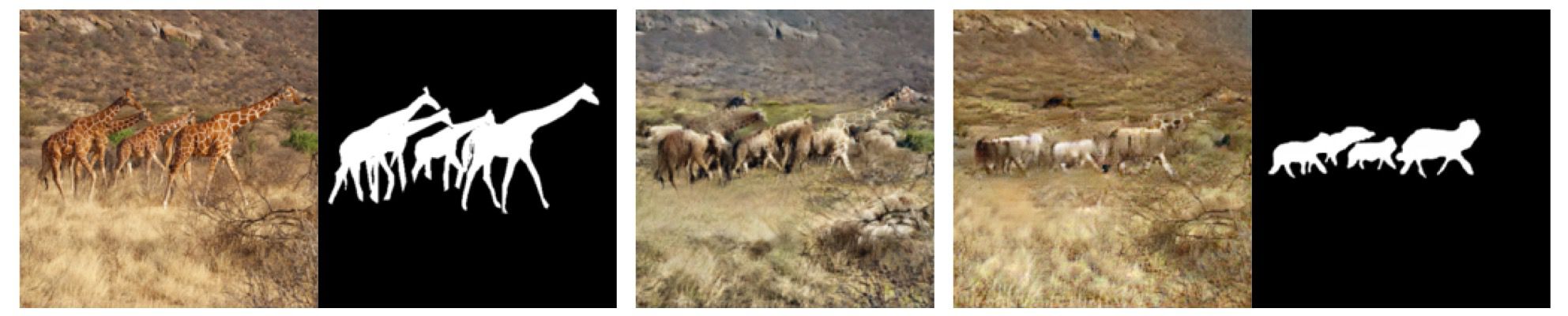}
	\caption{
	More translation results on COCO dataset (giraffe$\to$sheep).
	} \label{fig:more-coco-2}
\end{figure}

\newpage
\begin{figure}[H]
	\centering
	\includegraphics[width=\textwidth]{figure/appendix-label}
	\includegraphics[width=\textwidth]{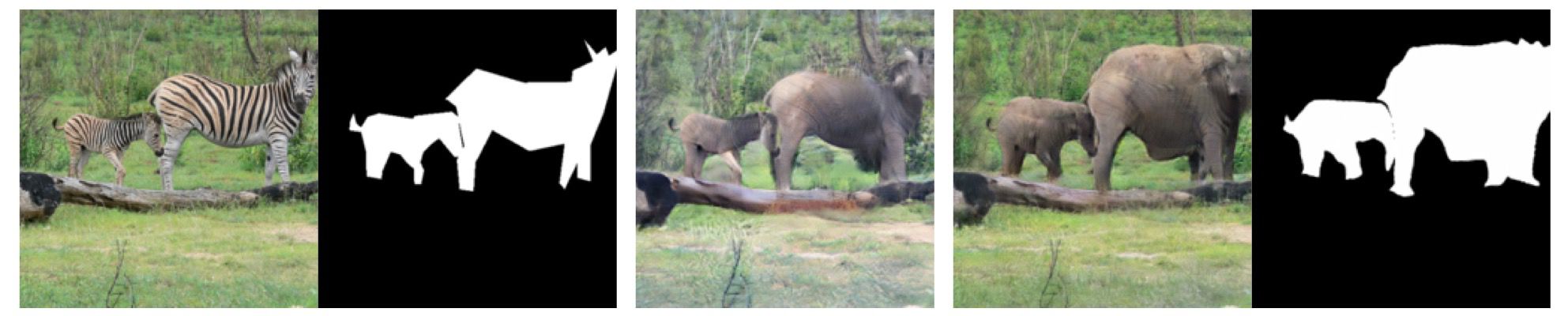}
    \includegraphics[width=\textwidth]{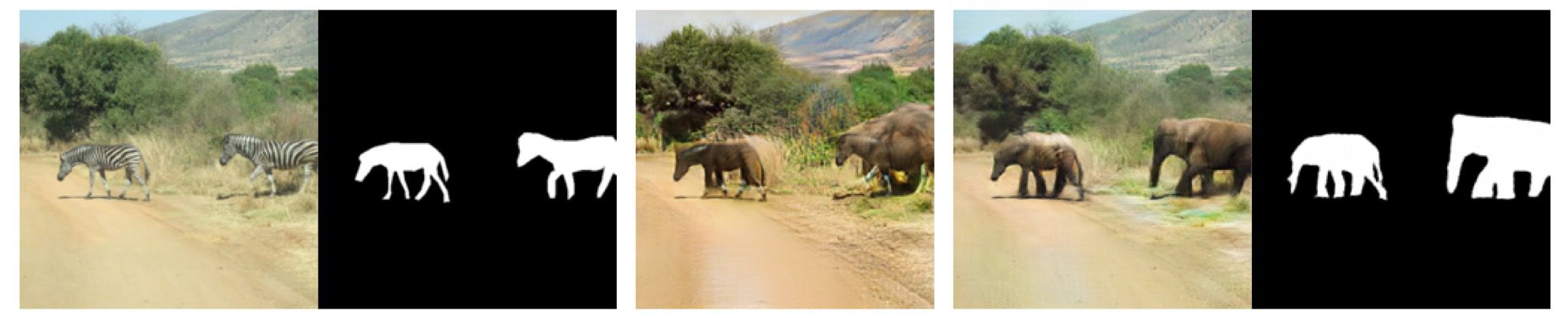}
	\caption{
	More translation results on COCO dataset (zebra$\to$elephant).
	} \label{fig:more-tasks-1}
\end{figure}

\begin{figure}[H]
	\centering
	\includegraphics[width=\textwidth]{figure/appendix-label}
	\includegraphics[width=\textwidth]{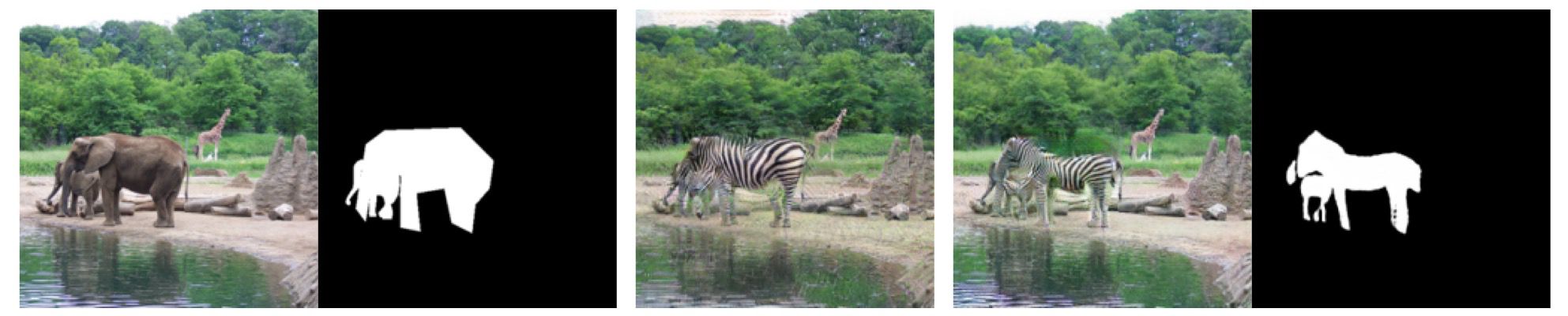}
    \includegraphics[width=\textwidth]{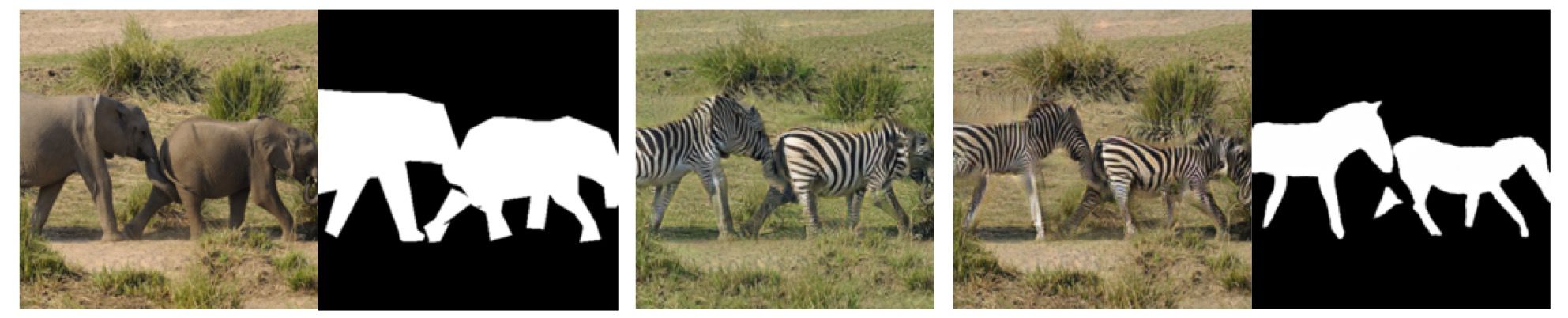}
	\caption{
	More translation results on COCO dataset (elephant$\to$zebra).
	} \label{fig:more-tasks-2}
\end{figure}

\begin{figure}[H]
	\centering
	\includegraphics[width=\textwidth]{figure/appendix-label}
	\includegraphics[width=\textwidth]{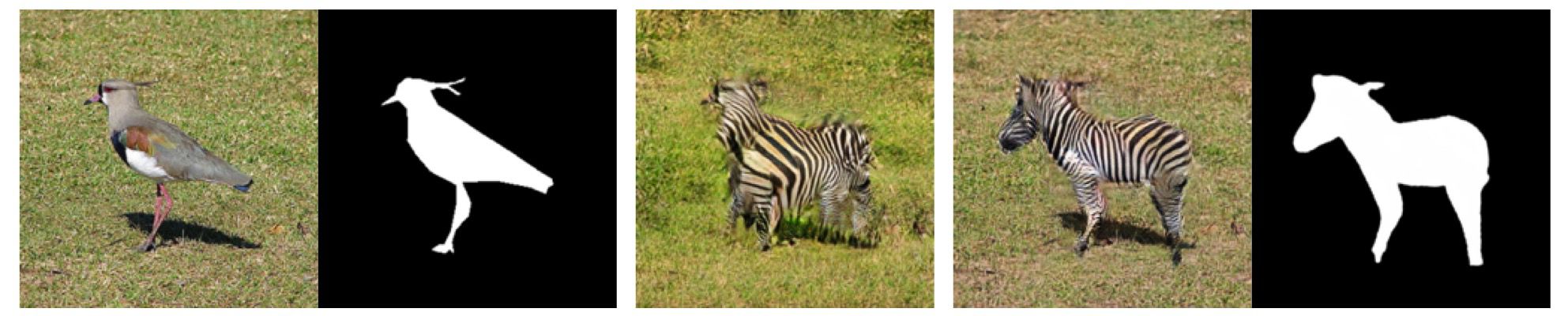}
    \includegraphics[width=\textwidth]{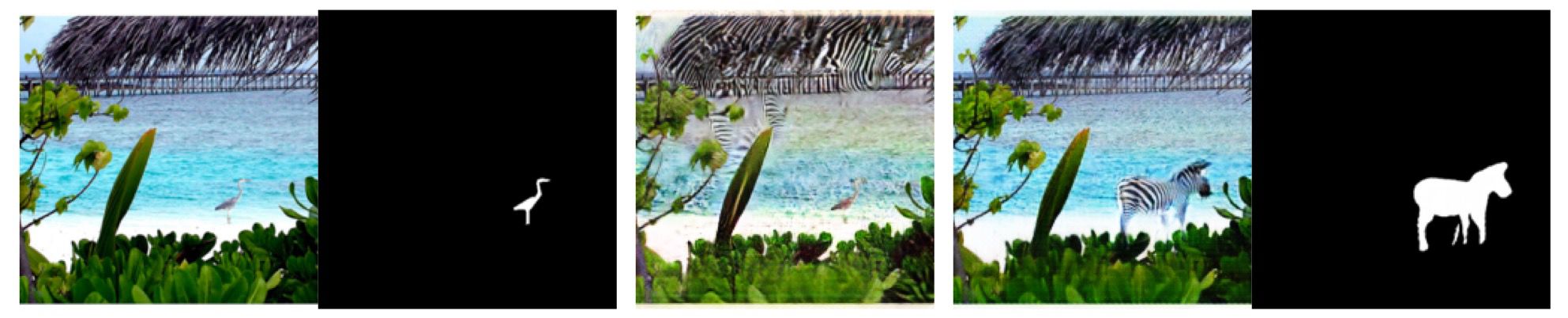}
	\caption{
	More translation results on COCO dataset (bird$\to$zebra).
	} \label{fig:more-tasks-3}
\end{figure}

\newpage
\begin{figure}[H]
	\centering
	\includegraphics[width=\textwidth]{figure/appendix-label}
	\includegraphics[width=\textwidth]{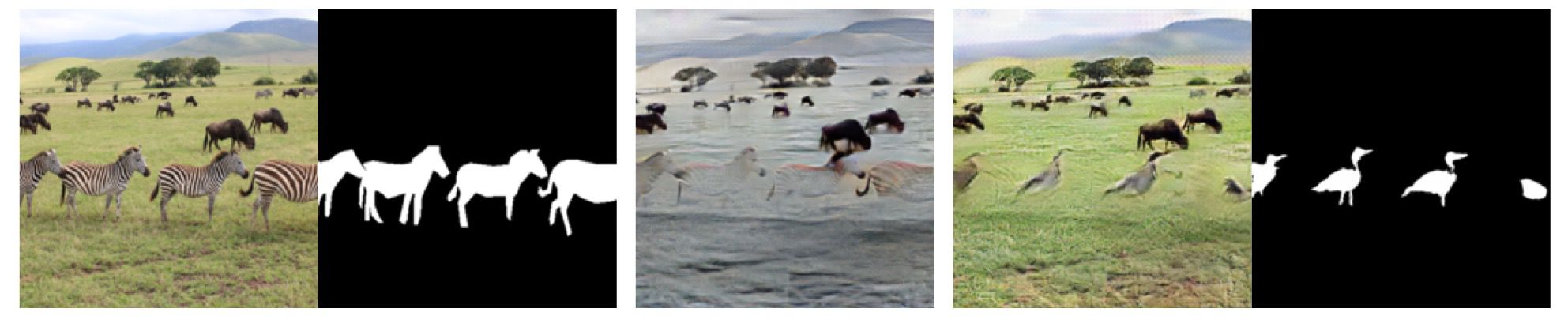}
    \includegraphics[width=\textwidth]{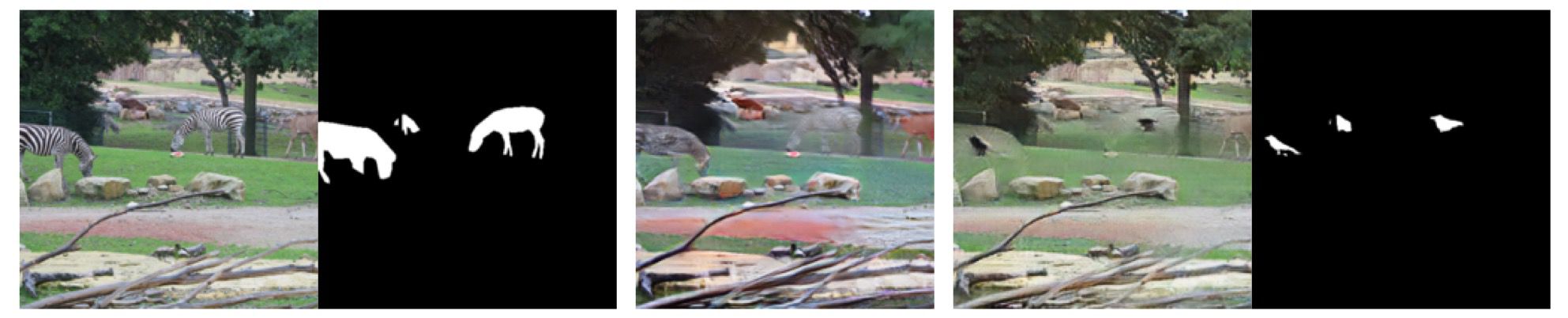}
	\caption{
	More translation results on COCO dataset (zebra$\to$bird).
	} \label{fig:more-tasks-4}
\end{figure}

\begin{figure}[H]
	\centering
	\includegraphics[width=\textwidth]{figure/appendix-label}
	\includegraphics[width=\textwidth]{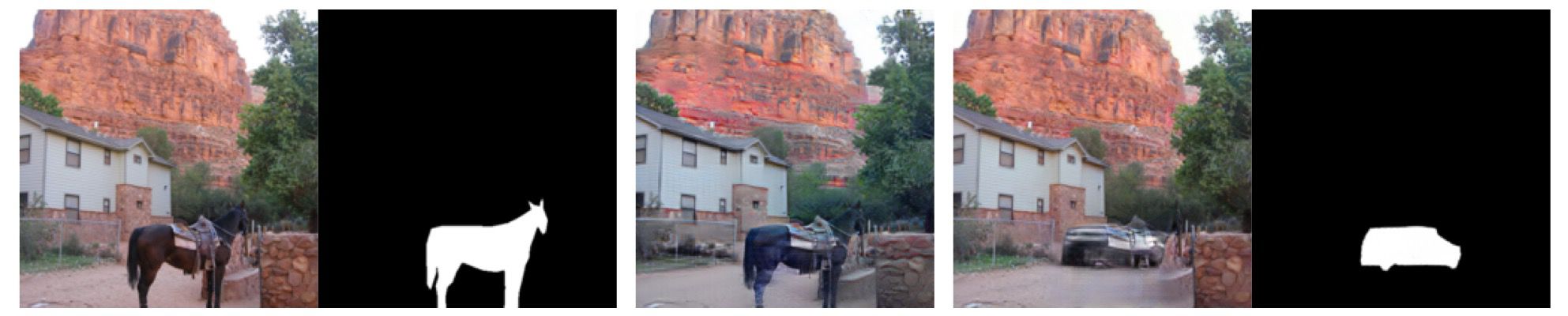}
    \includegraphics[width=\textwidth]{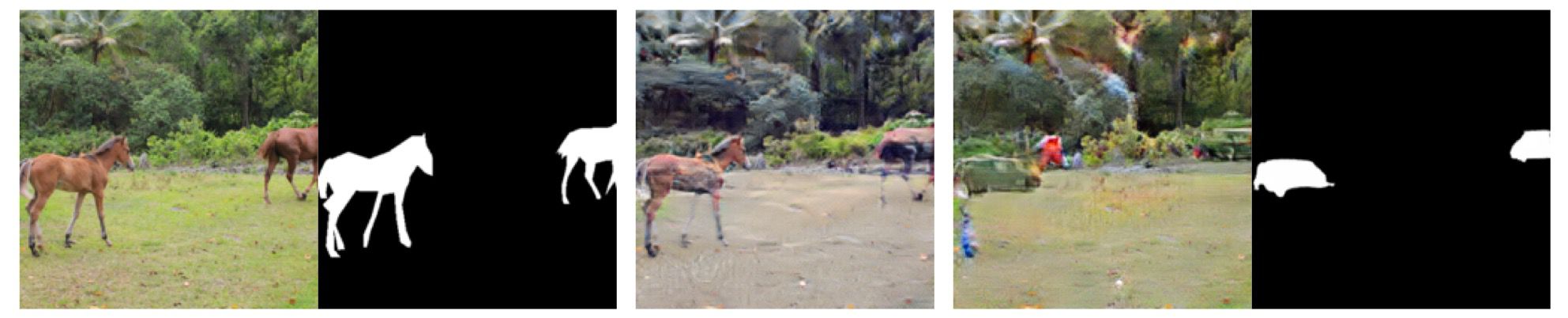}
	\caption{
	More translation results on COCO dataset (horse$\to$car).
	} \label{fig:more-tasks-5}
\end{figure}

\begin{figure}[H]
	\centering
	\includegraphics[width=\textwidth]{figure/appendix-label}
	\includegraphics[width=\textwidth]{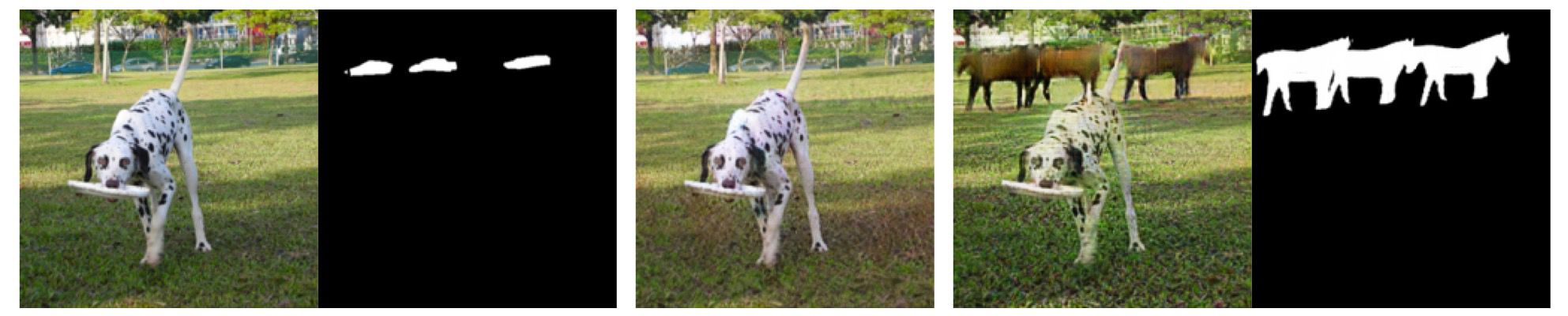}
    \includegraphics[width=\textwidth]{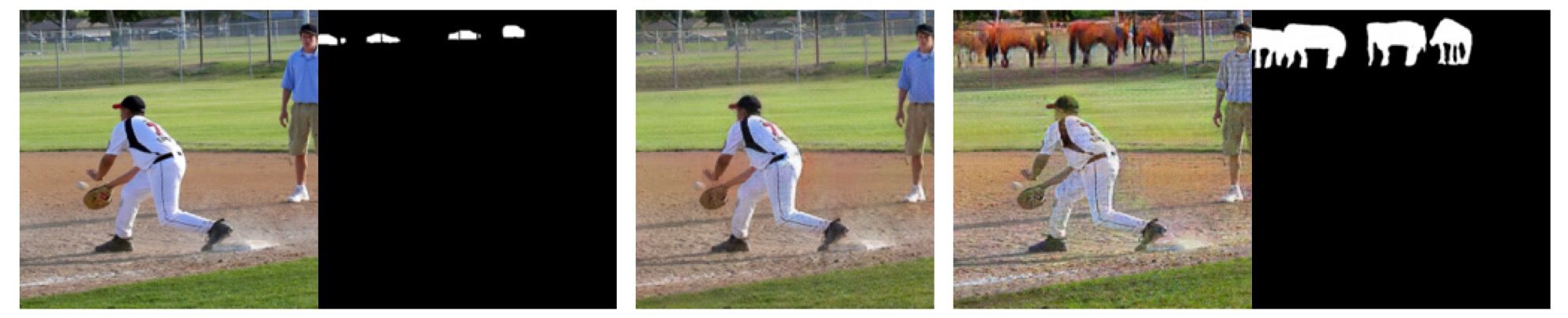}
	\caption{
	More translation results on COCO dataset (car$\to$horse).
	} \label{fig:more-tasks-6}
\end{figure}



\newpage
\section{More Comparisons with CycleGAN+Seg}
\label{sec:more-comparison}
\begin{figure}[h]
	\centering
	\includegraphics[width=\textwidth]{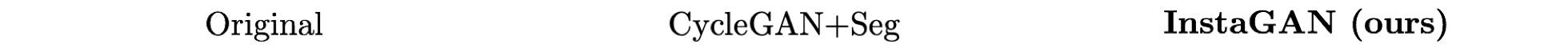}
	\includegraphics[width=\textwidth]{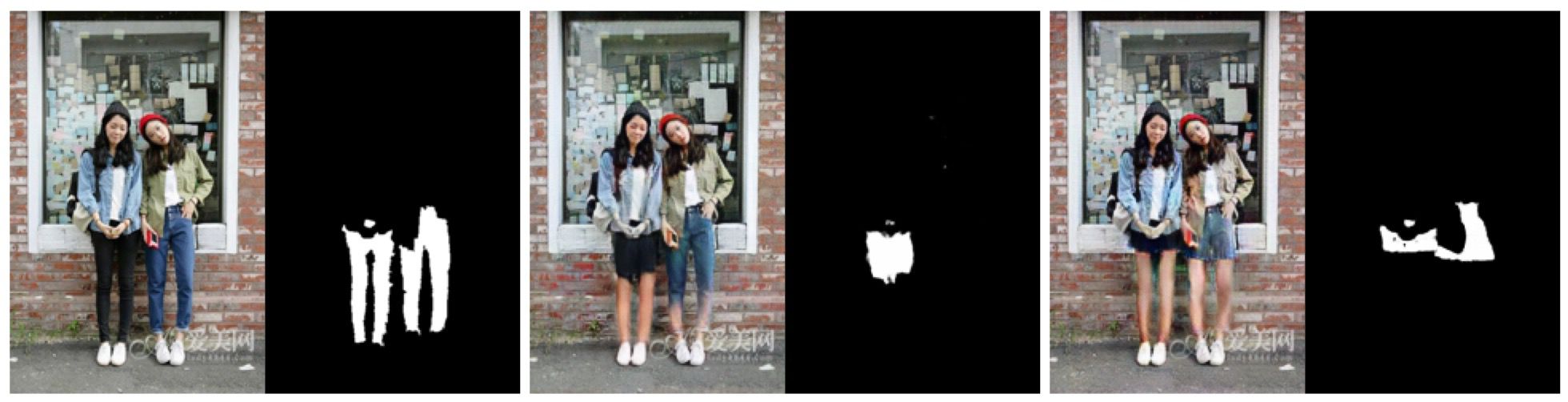}
	\includegraphics[width=\textwidth]{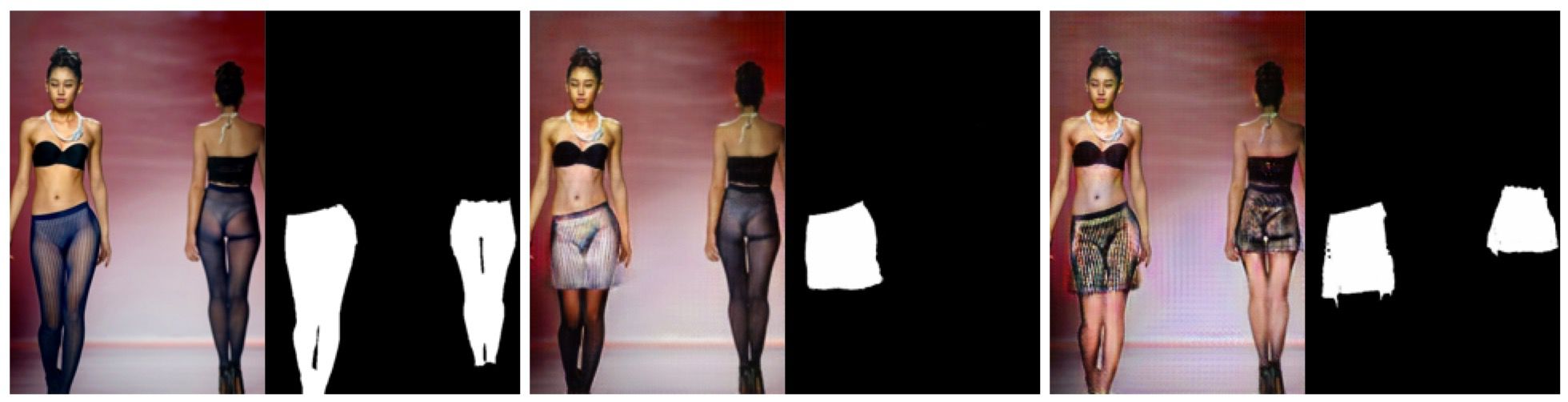}
	\includegraphics[width=\textwidth]{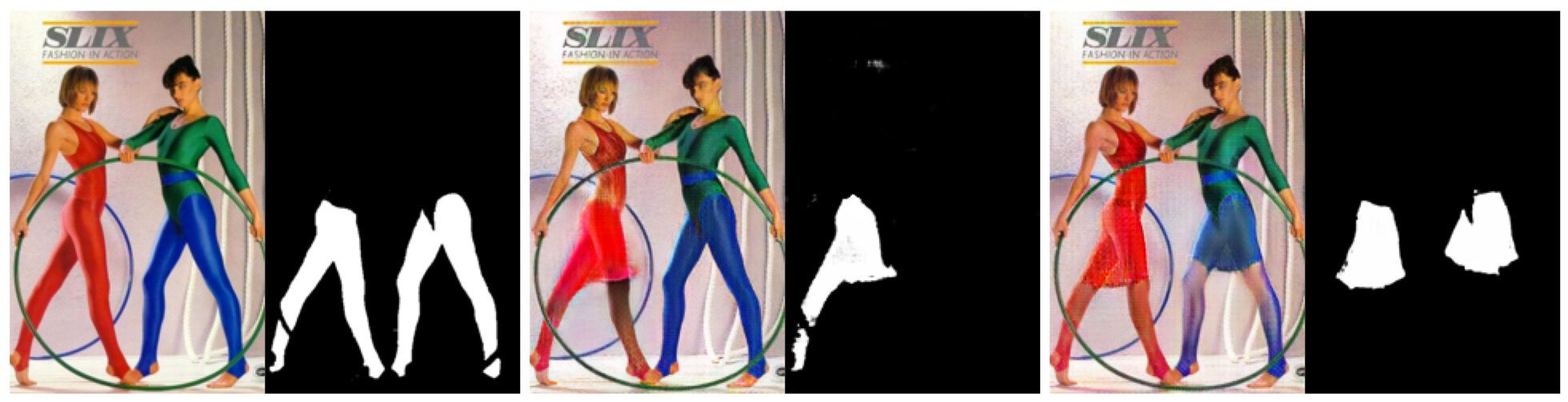}
	\includegraphics[width=\textwidth]{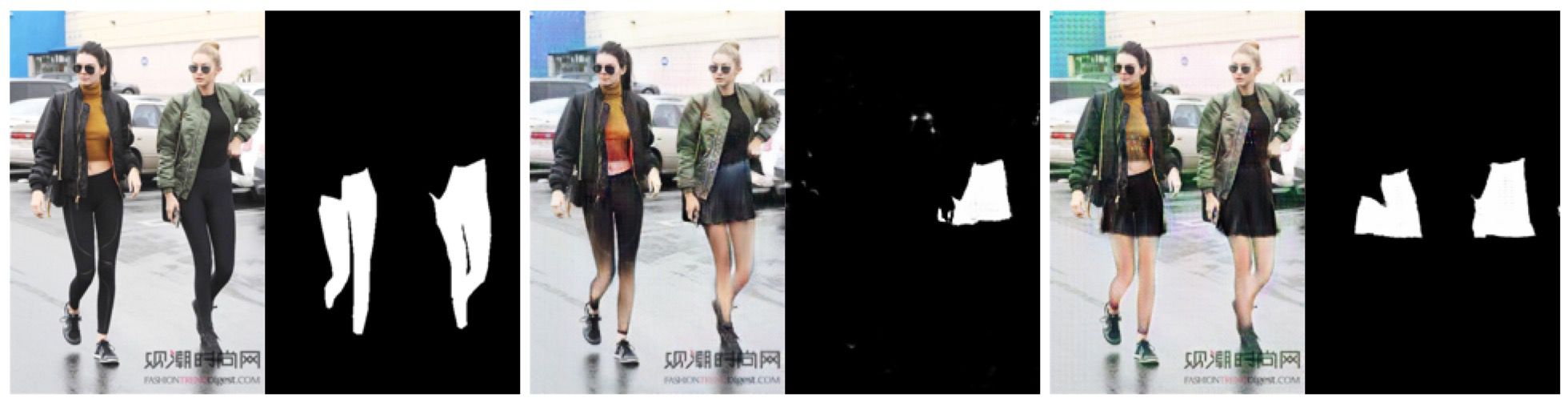}
	\includegraphics[width=\textwidth]{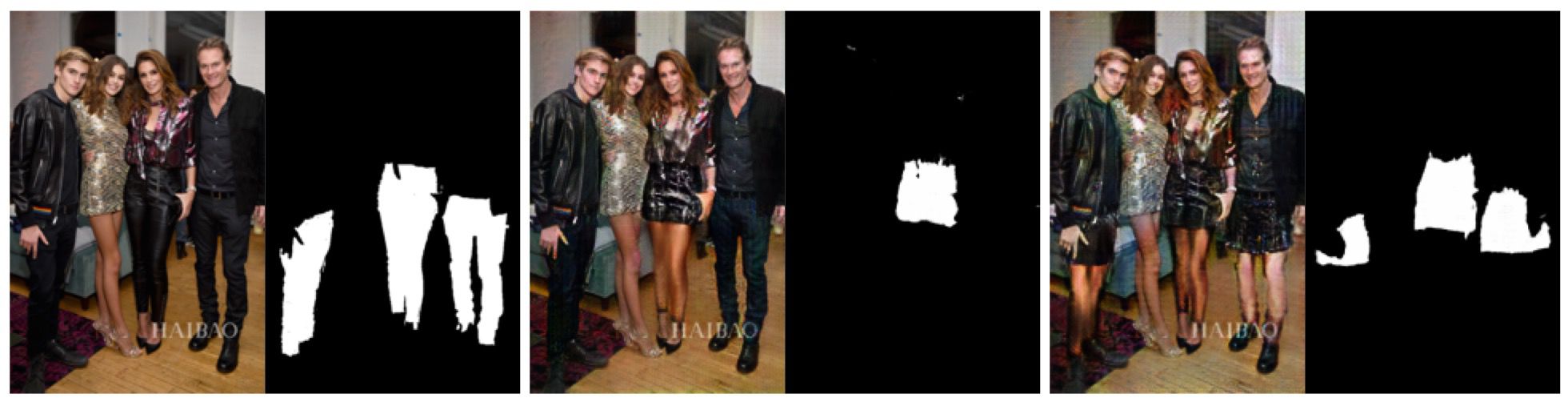}
	\caption{
	Comparisons with CycleGAN+Seg on MHP dataset (pants$\to$skirt).
	} \label{fig:more-cycleganseg-mhp-1}
\end{figure}

\begin{figure}[H]
	\centering
	\includegraphics[width=\textwidth]{figure/appendix-label-seg}
	\includegraphics[width=\textwidth]{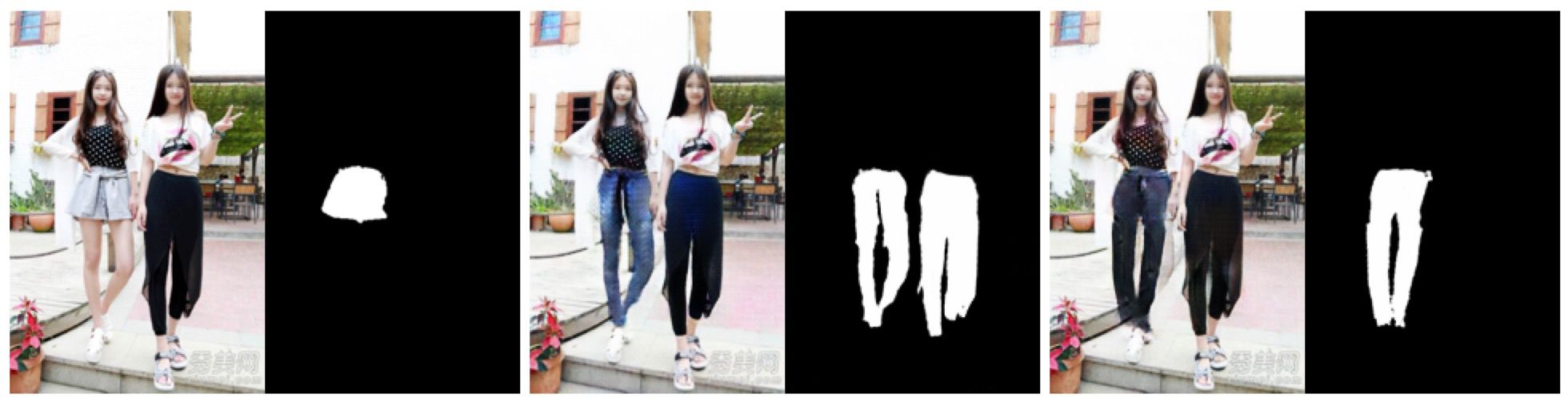}
	\includegraphics[width=\textwidth]{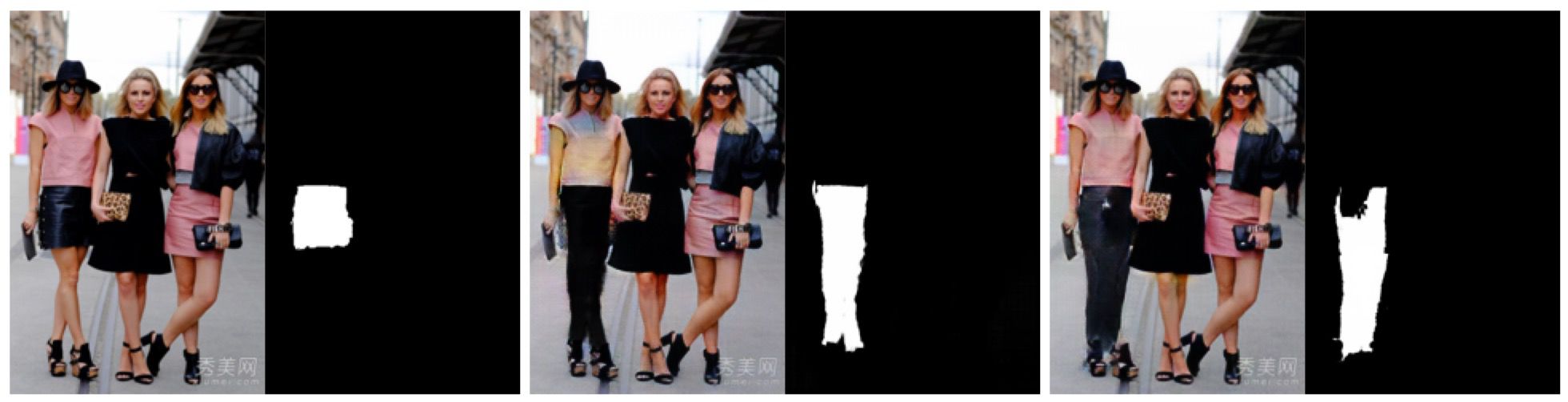}
	\includegraphics[width=\textwidth]{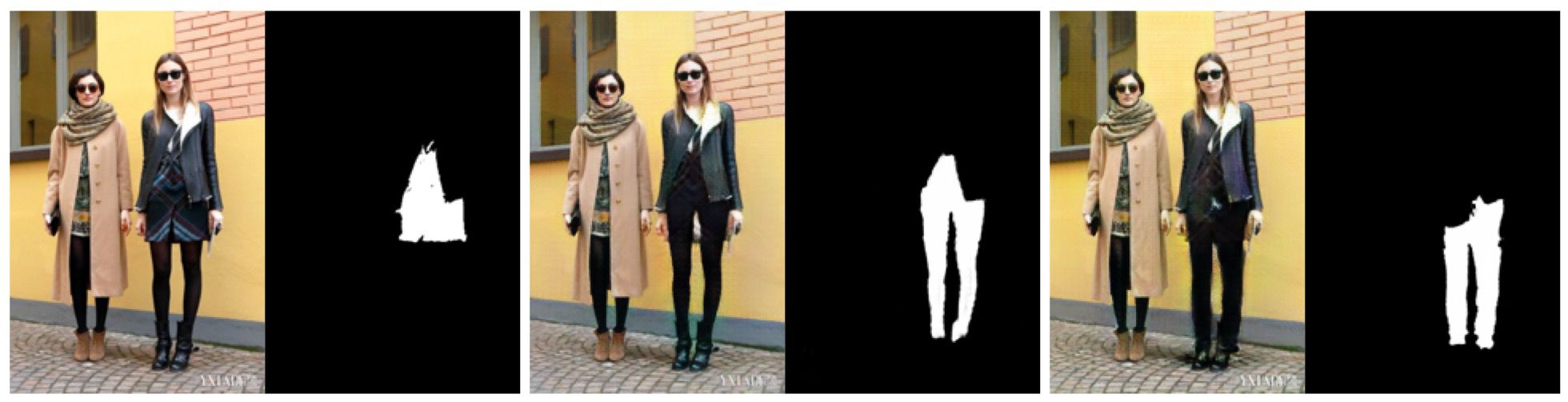}
	\includegraphics[width=\textwidth]{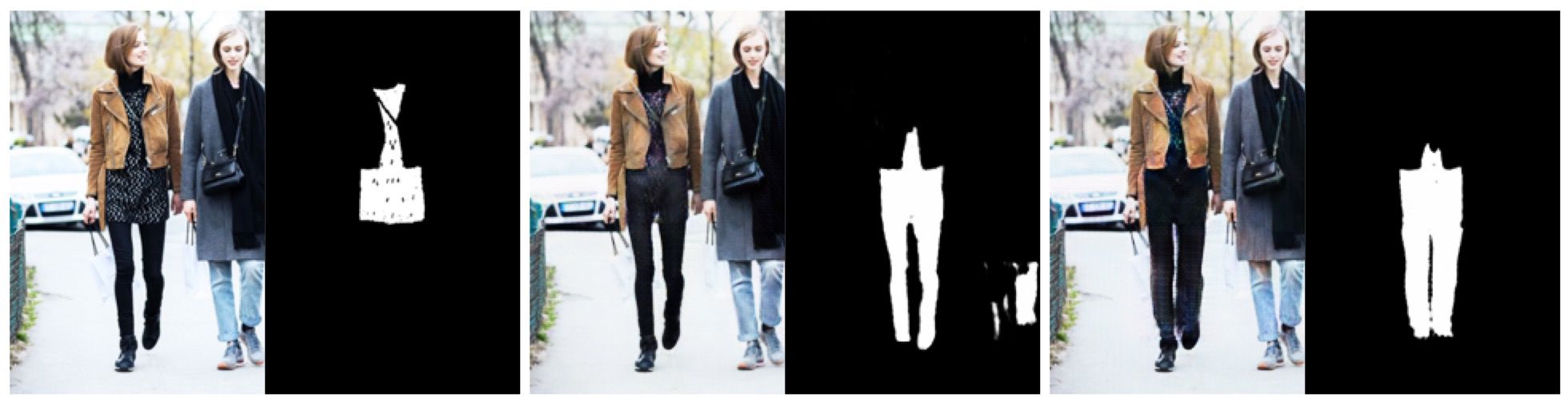}
	\includegraphics[width=\textwidth]{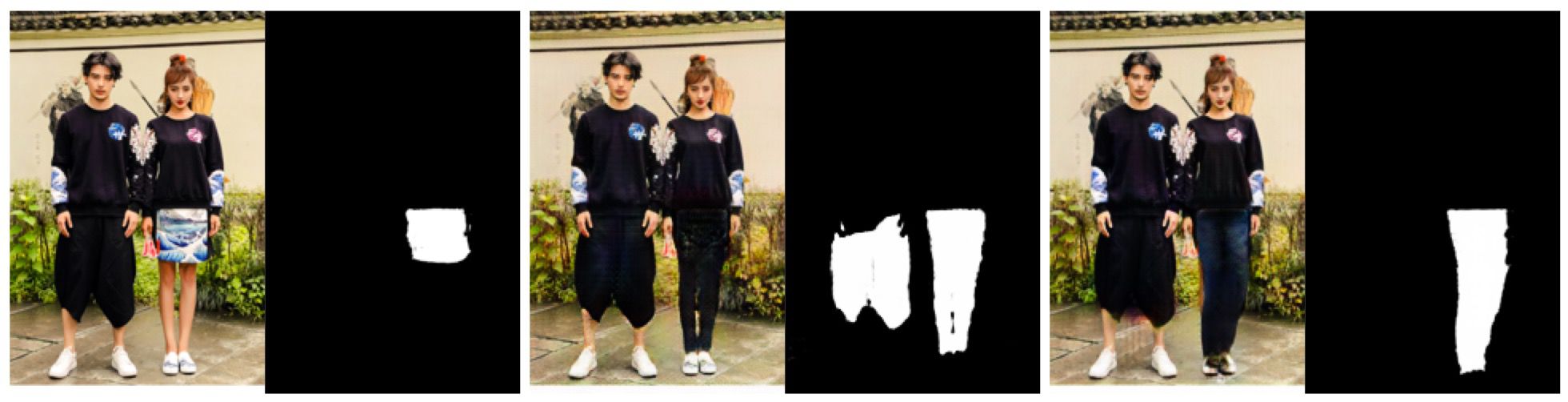}
	\caption{
	Comparisons with CycleGAN+Seg on MHP dataset (skirt$\to$pants).
	} \label{fig:more-cycleganseg-mhp-2}
\end{figure}

\begin{figure}[H]
	\centering
	\includegraphics[width=\textwidth]{figure/appendix-label-seg}
	\includegraphics[width=\textwidth]{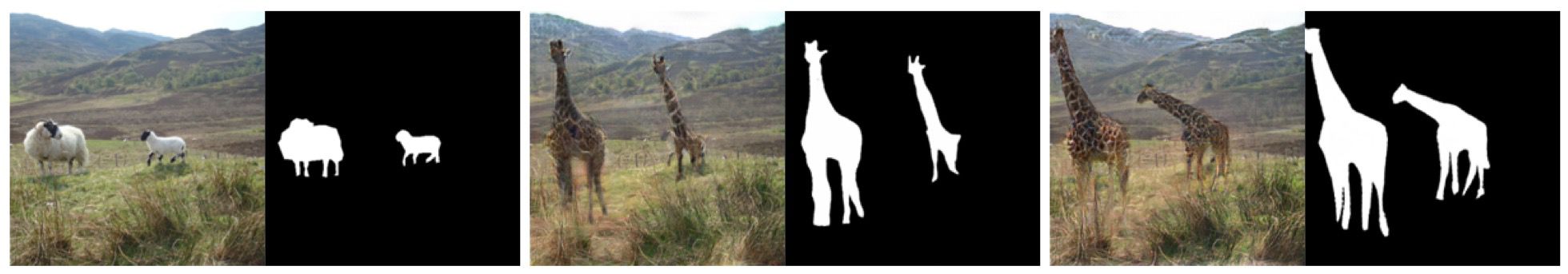}
    \includegraphics[width=\textwidth]{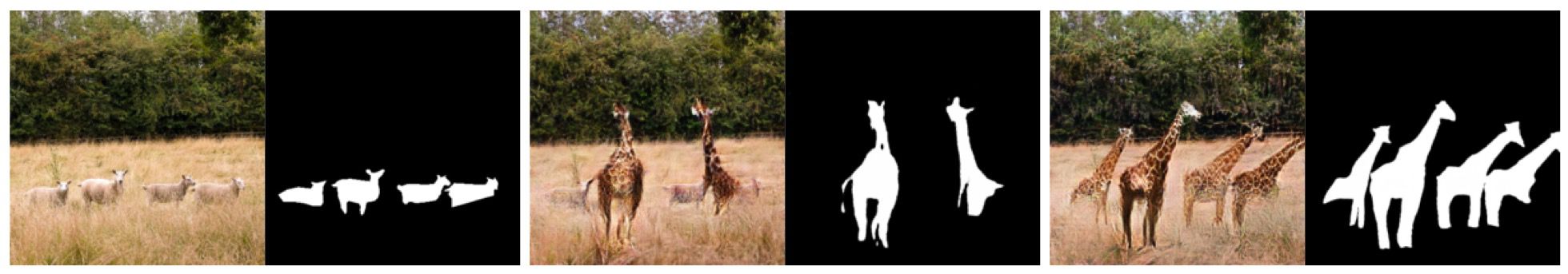}
    \includegraphics[width=\textwidth]{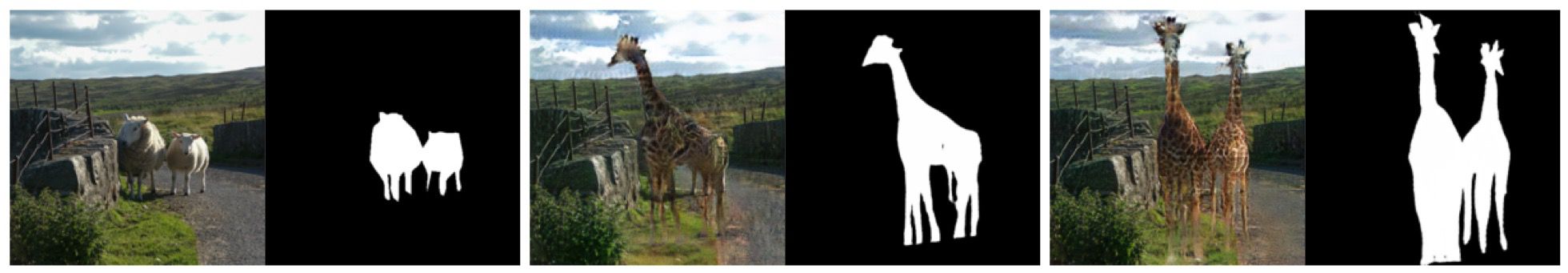}
    \includegraphics[width=\textwidth]{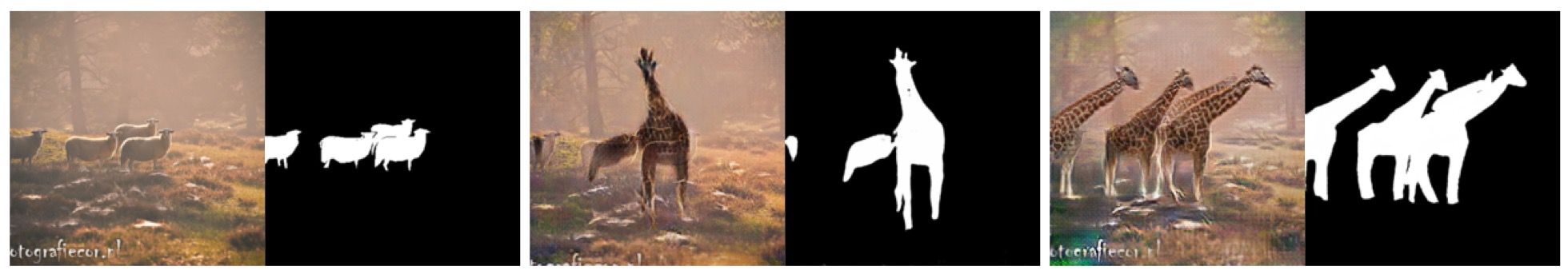}
    \includegraphics[width=\textwidth]{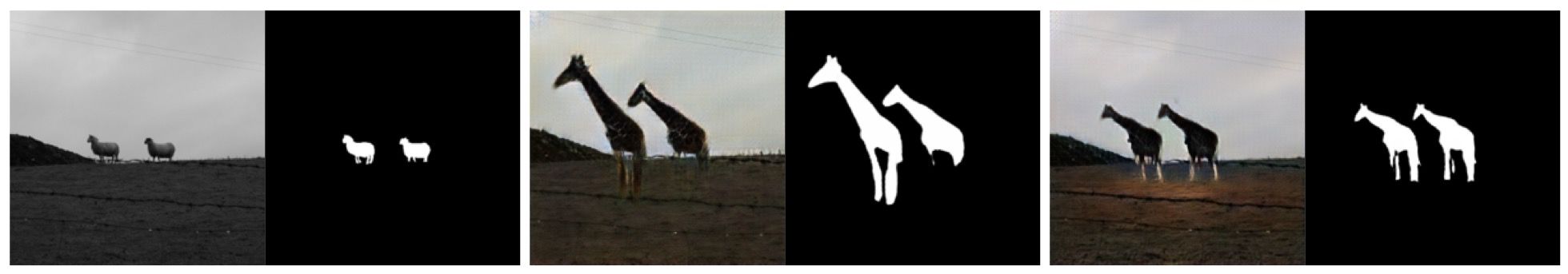}
    \includegraphics[width=\textwidth]{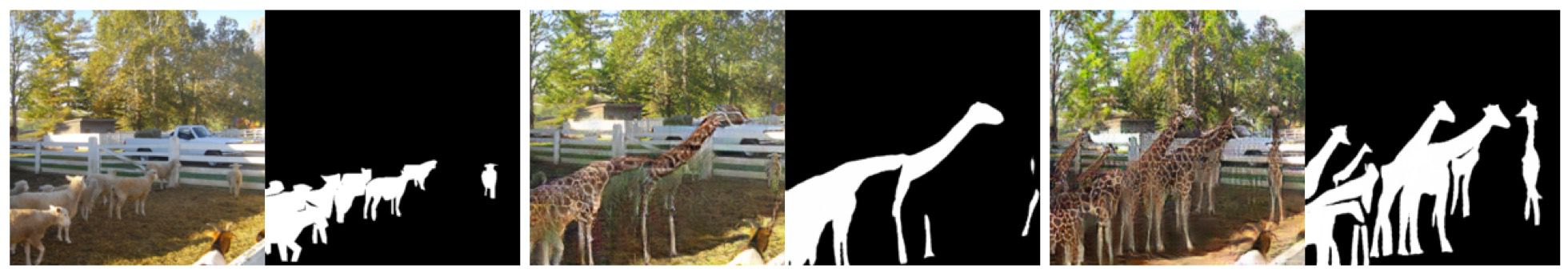}
    \includegraphics[width=\textwidth]{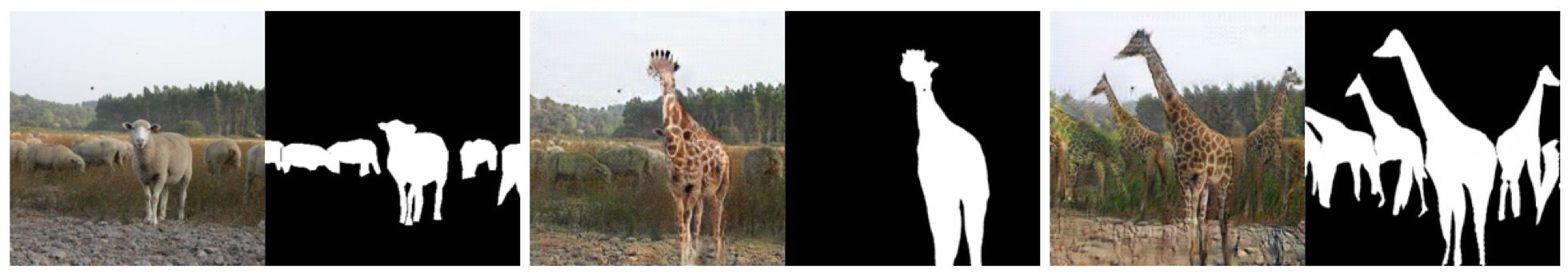}
	\caption{
	Comparisons with CycleGAN+Seg on COCO dataset (sheep$\to$giraffe).
	} \label{fig:more-cycleganseg-coco-1}
\end{figure}

\begin{figure}[H]
	\centering
	\includegraphics[width=\textwidth]{figure/appendix-label-seg}
	\includegraphics[width=\textwidth]{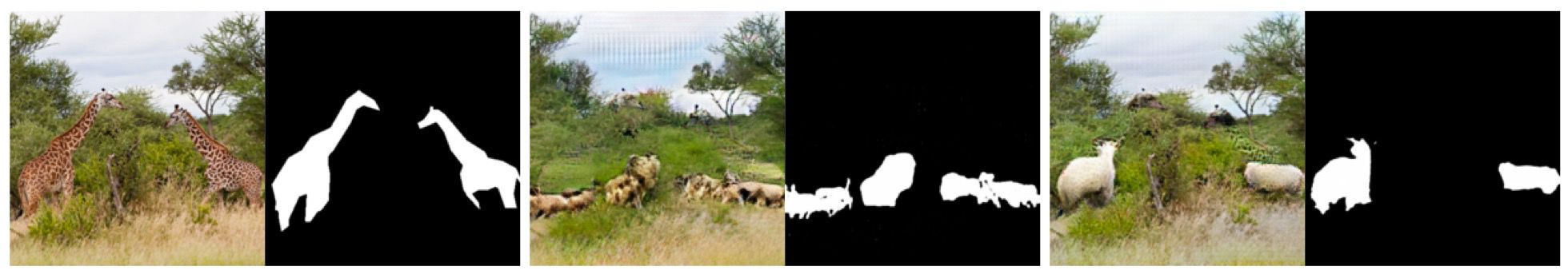}
    \includegraphics[width=\textwidth]{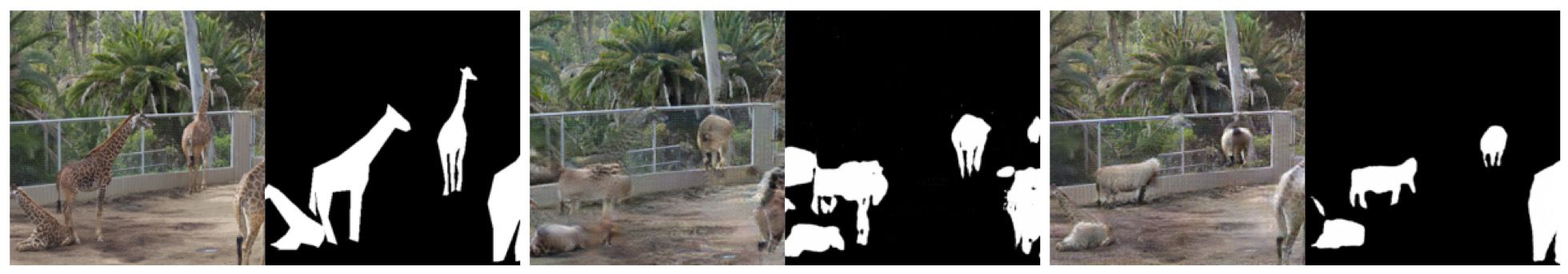}
    \includegraphics[width=\textwidth]{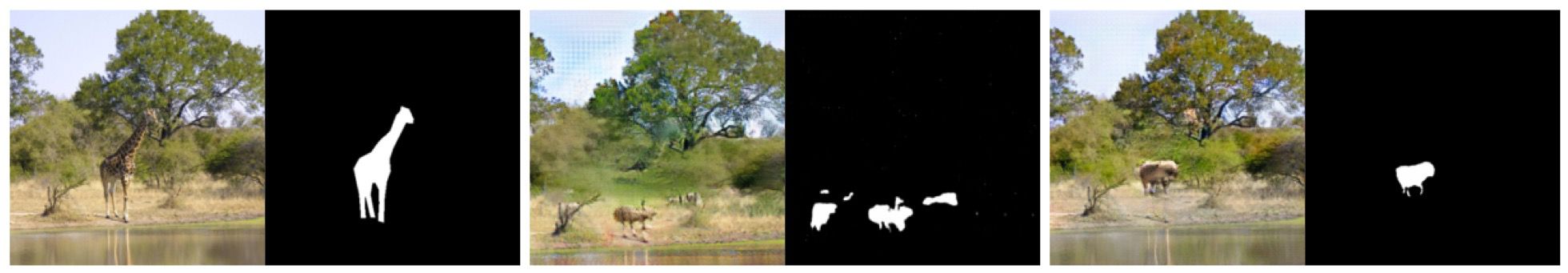}
    \includegraphics[width=\textwidth]{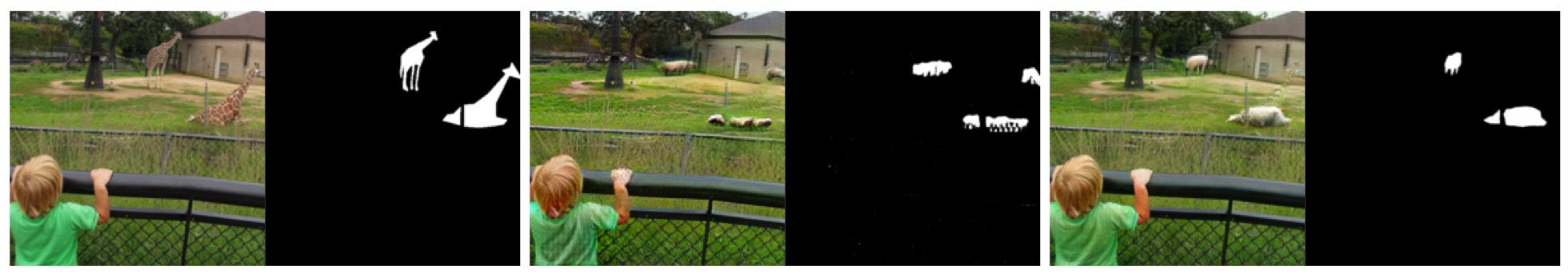}
    \includegraphics[width=\textwidth]{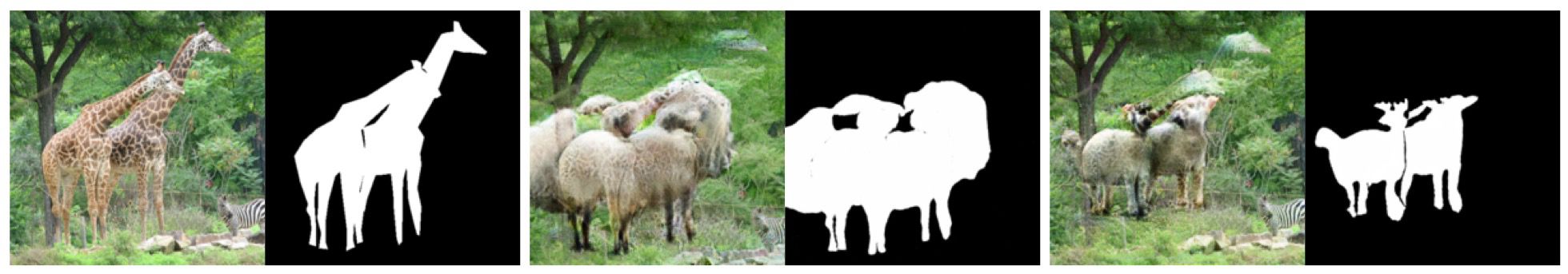}
    \includegraphics[width=\textwidth]{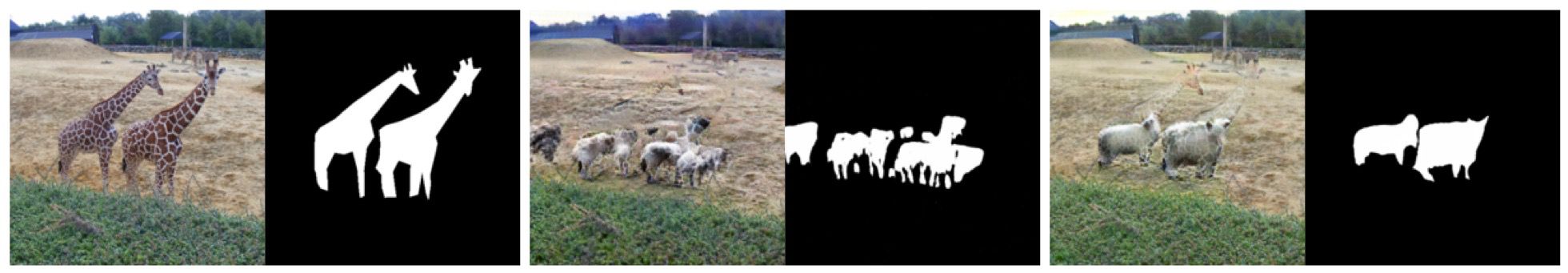}
    \includegraphics[width=\textwidth]{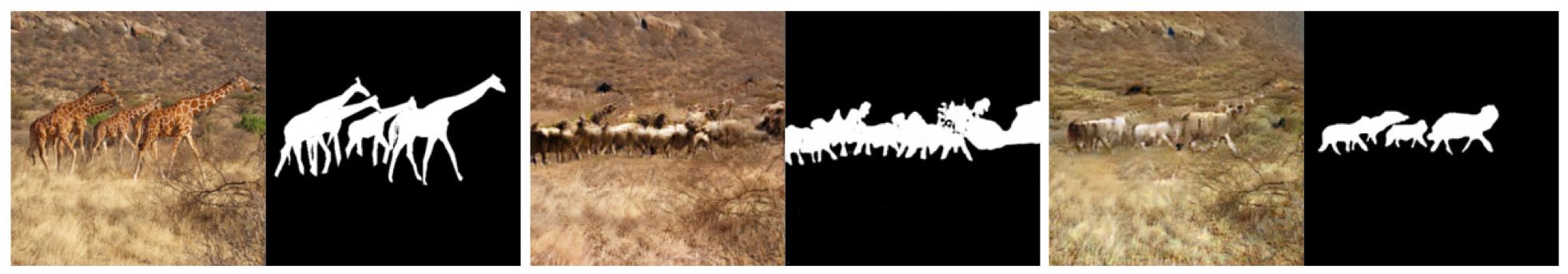}
	\caption{
	Comparisons with CycleGAN+Seg on COCO dataset (giraffe$\to$sheep).
	} \label{fig:more-cycleganseg-coco-2}
\end{figure}

To demonstrate the effectiveness of our method further,
we provide more comparison results with CycleGAN+Seg. 
Since CycleGAN+Seg translates all instances at once, it often
(a) fails to translate instances, or (b) merges multiple instances
(see Figure \ref{fig:more-cycleganseg-mhp-1} and \ref{fig:more-cycleganseg-coco-1}),
or (c) generates multiple instances from one instance
(see Figure \ref{fig:more-cycleganseg-mhp-2} and \ref{fig:more-cycleganseg-coco-2}).
On the other hand, our method does not have such issues due to its instance-aware nature.
In addition, since the unioned mask losses the original shape information,
our instance-aware method produces better shape results
(\textit{e.g.}, see row 1 of Figure \ref{fig:more-cycleganseg-coco-1}).

\newpage
\section{Generalization of Translated Masks}
\label{sec:nearest-neighbor}

\begin{figure}[H]
	\centering
	\includegraphics[width=\textwidth]{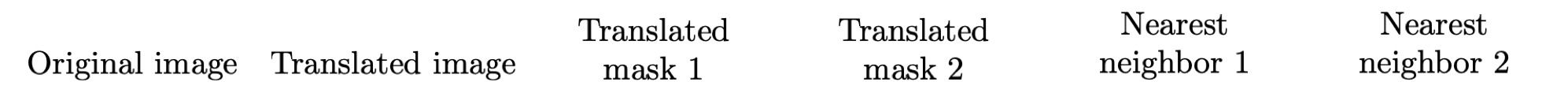}
	\includegraphics[width=\textwidth]{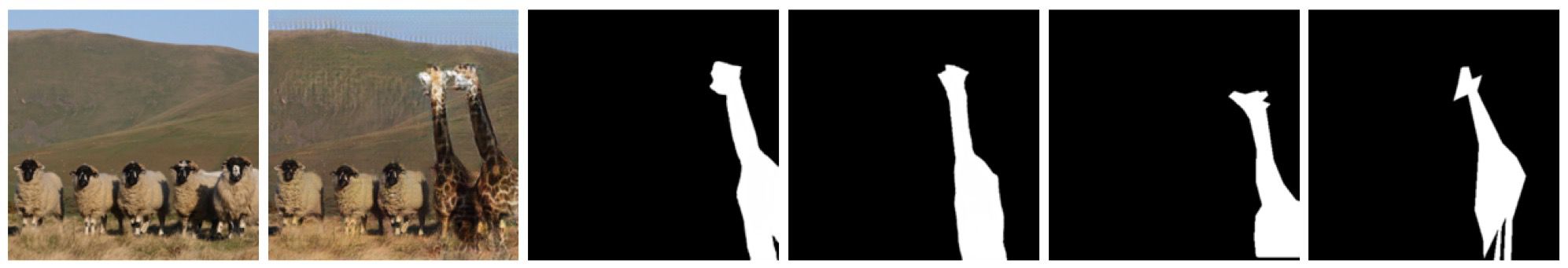}
    \includegraphics[width=\textwidth]{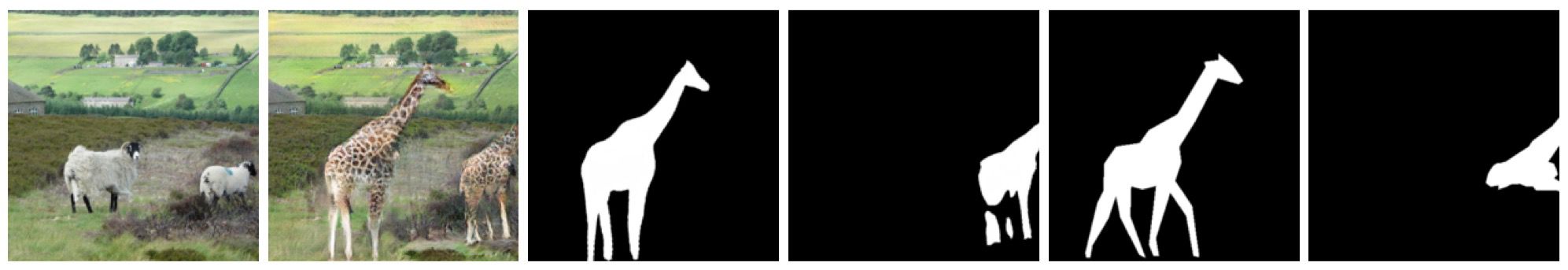}
	\caption{
	Nearest training neighbors of translated masks.
	} \label{fig:nearest-neighbor}
\end{figure}

To show that our model generalizes well,
we searched the nearest training neighbors (in $L_2$-norm) of translated target masks.
As reported in Figure \ref{fig:nearest-neighbor}, 
we observe that the translated masks (col 3,4) are often 
much different from the nearest neighbors (col 5,6).
This confirms that our model does not simply memorize training instance masks, but learns a mapping that generalizes for target instances.

\section{Translation Results of Crop \& Attach Baseline}

\begin{figure}[H]
	\centering
	\includegraphics[width=\textwidth]{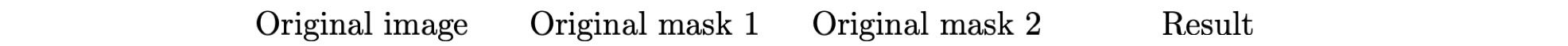}
	\includegraphics[width=\textwidth]{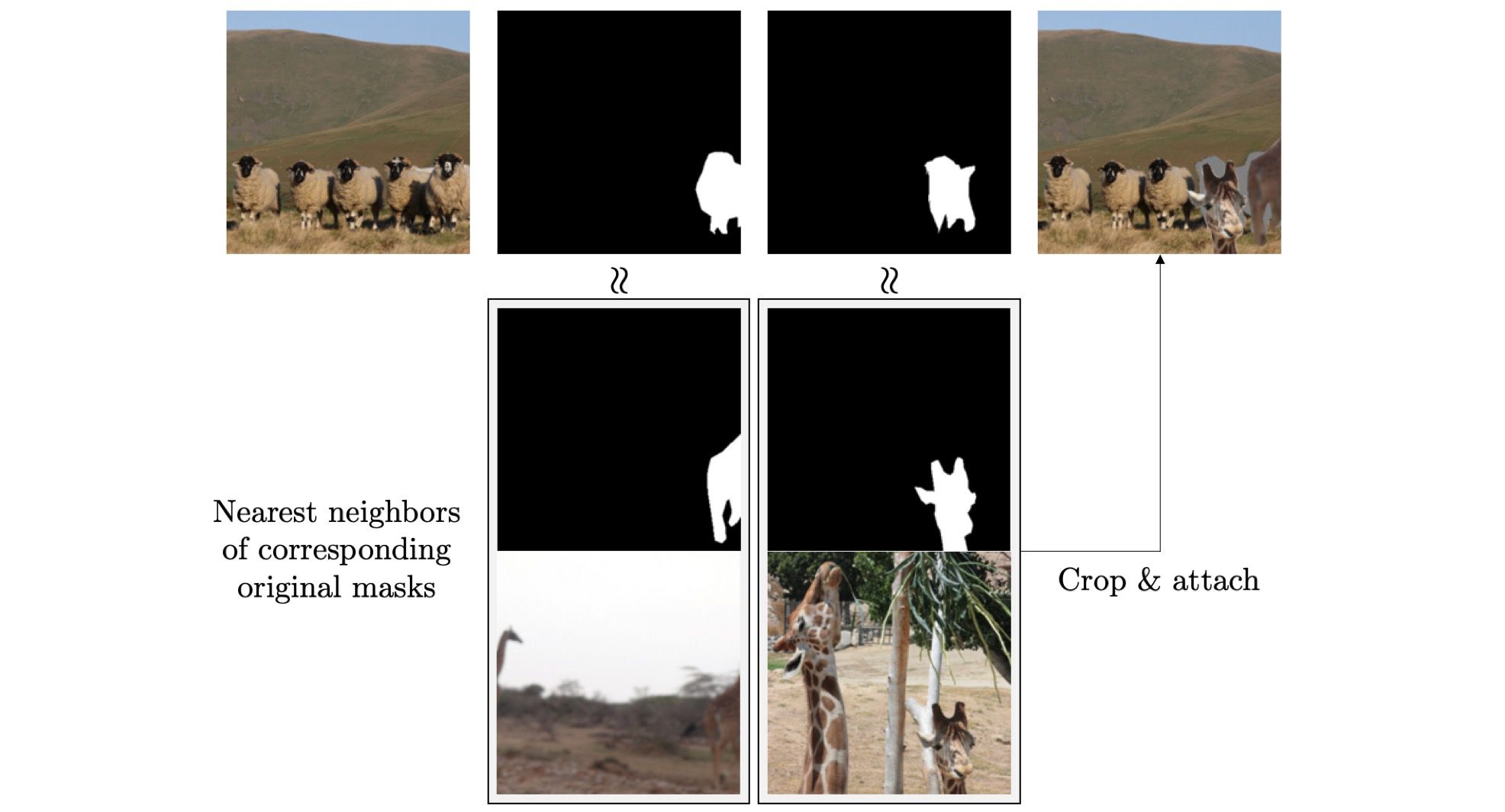}
	\caption{
	Translation results of crop \& attach baseline.
	} \label{fig:crop-and-attach}
\end{figure}

For interested readers, we also present the translation results of the simple crop \& attach baseline
in Figure \ref{fig:crop-and-attach},
that find the nearest neighbors of the original masks from target masks,
and crop \& attach the corresponding image to the original image.
Here, since the distance in pixel space (\textit{e.g.}, $L_2$-norm) obviously does not capture semantics,
the cropped instances do not fit with the original contexts as well.

\newpage
\section{Video Translation Results}

\begin{figure}[H]
	\centering
	\includegraphics[width=\textwidth]{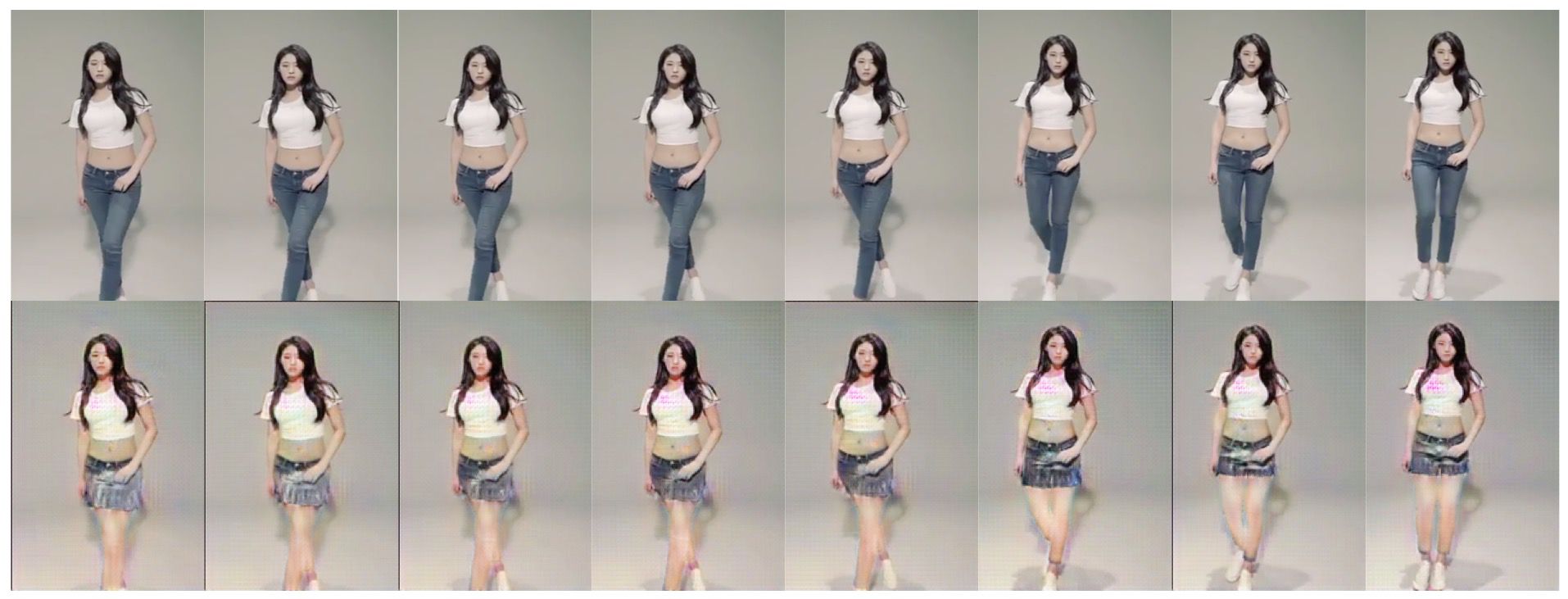}
	\caption{
	Original images (row 1) and translated results of our method (row 2) on a video searched from YouTube.
	We present translation results on successive eight frames for visualization.
	} \label{fig:video}
\end{figure}

For interested readers, we also present video translation results in Figure \ref{fig:video}.
Here, we use a predicted segmentation (generated by a pix2pix \citep{isola2017image} model as in
Figure \ref{fig:fashion-ccp-pred} and Figure \ref{fig:kpop}) for each frame.
Similar to CycleGAN, our method shows temporally coherent results, even though we did not used any explicit regularization.
One might design a more advanced version of our model utilizing temporal patterns
\textit{e.g.}, using the idea of Recycle-GAN \citep{bansal2018recycle} for video-to-video translation,
which we think is an interesting future direction to explore.

\newpage
\section{Reconstruction Results}

\begin{figure}[H]
	\centering
	\includegraphics[width=\textwidth]{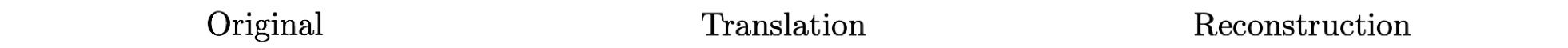}
	\includegraphics[width=\textwidth]{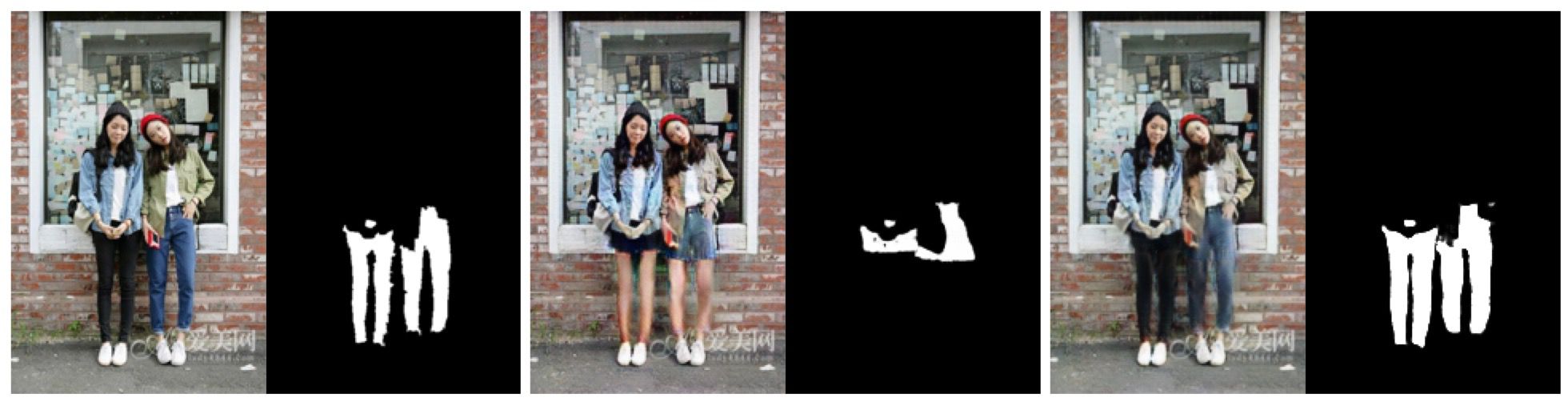}
    \includegraphics[width=\textwidth]{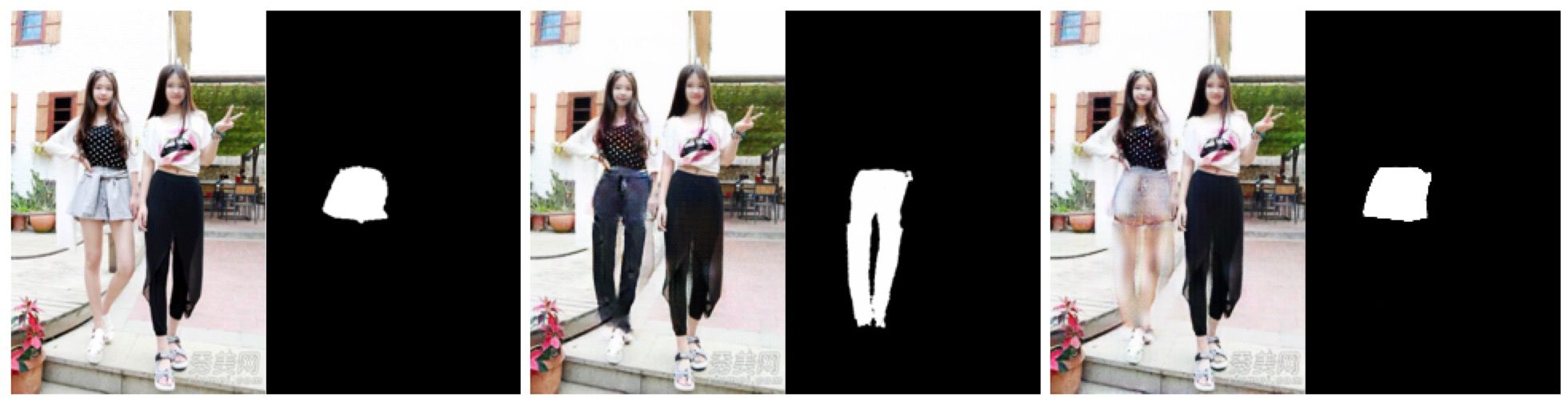}
    \includegraphics[width=\textwidth]{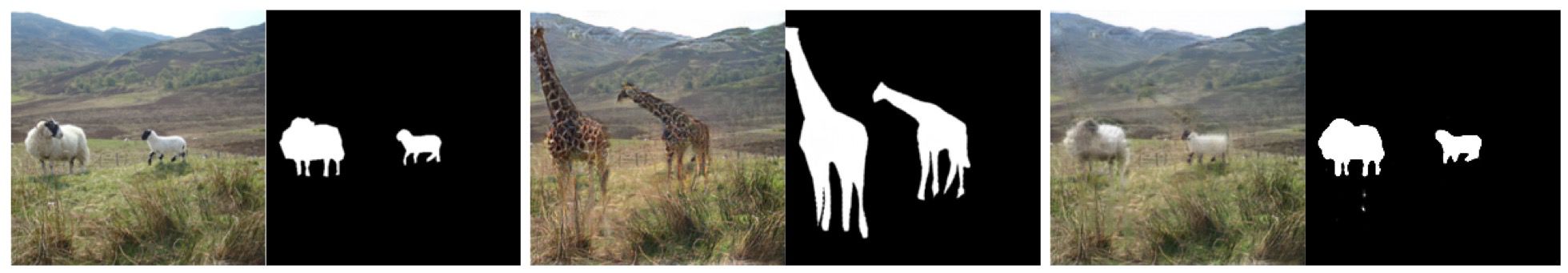}
    \includegraphics[width=\textwidth]{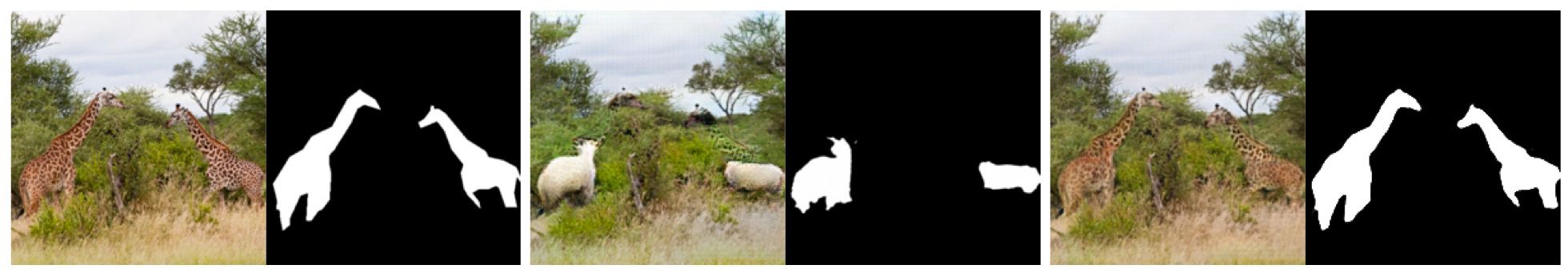}
	\caption{
	Translation and reconstruction results of our method.
	} \label{fig:recon}
\end{figure}

For interested readers, we also report the translation and reconstruction results of our method in Figure
\ref{fig:recon}. One can observe that 
our method shows good reconstruction results while showing good translation results.
This implies that our translated results preserve the original context well.


\end{document}